\documentclass[10pt]{article} % For LaTeX2e
\usepackage[accepted]{tmlr}
\usepackage{amsmath,amsfonts,bm}

\def\eqref#1{equation~\ref{#1}}
\def\1{\bm{1}}

\DeclareMathAlphabet{\mathsfit}{\encodingdefault}{\sfdefault}{m}{sl}
\SetMathAlphabet{\mathsfit}{bold}{\encodingdefault}{\sfdefault}{bx}{n}

\usepackage{hyperref}       % hyperlinks
\usepackage{url}            % simple URL typesetting
\usepackage{booktabs}       % professional-quality tables
\usepackage{amsfonts}       % blackboard math symbols
\usepackage{nicefrac}       % compact symbols for 1/2, etc.
\usepackage{microtype}      % microtypography
\usepackage{xcolor}         % colors
\usepackage{pifont}
\usepackage{graphicx}
\usepackage{multirow}
\usepackage{amsmath}
\usepackage{tabularx}
\usepackage{forest}
\usetikzlibrary{shadows}
\usepackage{cleveref}
\usepackage{booktabs}
\usepackage{wrapfig}
\usepackage{xspace}
\usepackage{subcaption}

\definecolor{lightcoral}{rgb}{0.94, 0.5, 0.5}
\definecolor{lightgreen}{rgb}{0.56, 0.93, 0.56}
\definecolor{harvestgold}{rgb}{0.85, 0.57, 0.0}
\definecolor{brightlavender}{rgb}{0.75, 0.58, 0.89}
\definecolor{capri}{rgb}{0.0, 0.75, 1.0}
\definecolor{carminepink}{rgb}{0.92, 0.3, 0.26}
\definecolor{celadon}{rgb}{0.67, 0.88, 0.69}
\definecolor{darkpastelgreen}{rgb}{0.01, 0.75, 0.24}

\newcommand{\paex}{$\mathcal{E}$\xspace}
\newcommand{\modelp}{\textbf{Saving \paex into Model Parameters}\xspace}
\newcommand{\memory}{\textbf{Saving \paex into Explicit Memory}\xspace}
\newcommand{\rag}{\textbf{Saving \paex into Knowledge Bases for Retrieval}\xspace}
\newcommand{\lc}{\textbf{Saving \paex into Raw Text and Process All}\xspace}

\NewDocumentCommand{\heng}
{ mO{} }{\textcolor{red}{\textsuperscript{\textit{Heng}}\textsf{\textbf{\small[#1]}}}}

\newcommand{\yes}{\textcolor{green}{\ding{51}}}
\newcommand{\no}{\textcolor{red}{\ding{55}}}
\newcommand{\wy}[1]{{\color{black}{#1}}}

\title{Towards LifeSpan Cognitive Systems}

\author{\name Yu Wang$^1$\thanks{Y. Wang, C. Han, T. Wu and X. He contribute equally. Correspondence to \texttt{yuw164@ucsd.edu}.}, Chi Han$^{2*}$, Tongtong Wu$^{3*}$, Xiaoxin He$^{4*}$, \\
\textbf{Wangchunshu Zhou$^5$, Nafis Sadeq$^1$, Xiusi Chen$^{2,6}$, Zexue He$^{1,7}$}, \\
\textbf{Wei Wang$^6$, Gholamreza Haffari$^3$, Heng Ji$^2$, Julian McAuley$^1$}\\
\addr $^1$UCSD, $^2$UIUC, $^3$Monash, $^4$NUS, $^5$AIWaves, $^6$UCLA, $^7$MIT-IBM\\}

\def\month{02}  % Insert correct month for camera-ready version
\def\year{2025} % Insert correct year for camera-ready version
\def\openreview{\url{https://openreview.net/forum?id=LZ9FmeFeLV}} % Insert correct link to OpenReview for camera-ready version

\begin{document}

\maketitle

\begin{abstract}
Building a human-like system that continuously interacts with complex environments—whether simulated digital worlds or human society—presents several key challenges. Central to this is enabling continuous, high-frequency interactions, where the interactions are termed experiences. We refer to this envisioned system as the \textbf{LifeSpan Cognitive System (LSCS)}.
A critical feature of LSCS is its ability to engage in incremental and rapid updates while retaining and accurately recalling past experiences. \wy{In this paper we focus on the domain of Large Language Models (LLMs), where we} identify two major challenges: (1) Abstraction and Experience Merging, and (2) Long-term Retention with Accurate Recall. These properties are essential for storing new experiences, organizing past experiences, and responding to the environment in ways that leverage relevant historical data.
Unlike language models with continual learning, which typically rely on large corpora for fine-tuning and focus on improving performance within specific domains or tasks, LSCS must rapidly and incrementally update with new information from its environment at a high frequency.
Existing technologies with the potential of solving the above two major challenges can be classified into four classes based on a conceptual metric called \textbf{Storage Complexity}, which measures the relative space required to store past experiences. Each of these four classes of technologies has its own strengths and limitations while we argue none of them alone can achieve LSCS alone. To this end, we propose a potential instantiation for LSCS that can integrate all four classes of technologies.
The new instantiation, serving as a conjecture, operates through two core processes: Absorbing Experiences and Generating Responses. 
\end{abstract}
\vspace{-10pt}

\section{Introduction}
\label{sec:introduction}
% \vspace{-5pt}
Imagine an environment such as a simulated digital world~\citep{park2023generative}, a virtual world like Minecraft~\citep{voyager}, the Marvel Universe depicted in movies, or even the complexities of human society. Developing an intelligent system capable of engaging with such environments—interacting with its surroundings, absorbing information from experiences, self-evolving based on feedback~\citep{zhou2024symbolic}, and living through an entire lifespan, either as a real or virtual entity—remains a significant challenge. When the system inhabits the environment, it interacts with everything around it, including humans, objects, and other elements of the environment. We define these interactions as the system’s \textbf{experiences}, encompassing all of the system’s behaviors and the feedback it receives from the environment—whether visual, physical, linguistic, or across any other modality.
This system’s cognitive capabilities include the ability to perceive and interpret its surroundings, retain and recall information through memory, learn and adapt from feedback, and engage in reasoning and decision-making processes. These cognitive functions enable the system to navigate its environment intelligently and evolve over time. We denote the system with the above capabilities as the \textbf{LifeSpan Cognitive System (LSCS)}.
\wy{A real-world example of LSCS is a long-lived, autonomous AI agent—such as a virtual companion or an agent operating within a simulated AI civilization. Over time, it would continuously interact with its environment, accumulate experiences, and refine its internal models. These interactions might span years, requiring the agent to recall past events and lessons learned weeks, months, or years earlier. 
In this paper, we mainly focus on constructing LSCS based on large language models (LLMs), as LLMs are currently the most promising direction to achieve human-like systems. Moreover, we specifically focus on text domain as most technologies discussed in this paper are constrained in text domain. 
}

\begin{figure*}[t]
\centering
\includegraphics[width=0.9\linewidth]{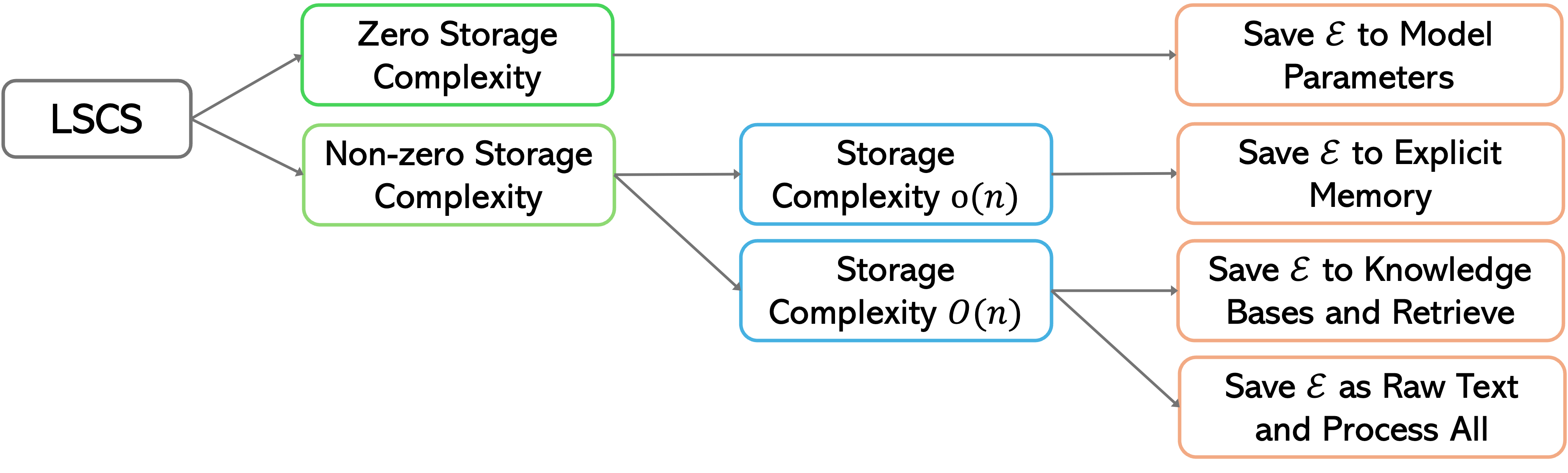}
\caption{
The technologies for constructing Lifespan Cognitive Systems (LSCS) can be broadly categorized into two principal approaches based on storage complexity. The first approach, characterized by zero storage complexity, involves 
\modelp. The second approach, which involves non-zero storage complexity, is subdivided into methods with storage complexities of $o(n)$ and $O(n)$. The $o(n)$ methods indicate \memory. The $O(n)$ methods are further classified based on whether the language model in the system processes the entire stored text. Suppose an additional retriever is adopted and the language model only accesses a snippet of the stored experiences. In that case, it falls under the category of \rag. The final category encompasses methods where the entire context is input into the model, classified as \lc.
} 
\label{fig:categorization}
\end{figure*}

\begin{table*}
\centering
\begin{tabular}{cccccccc}
    \toprule
    Where to store & Sub-Cat & Abs \& Ex-Mg & Long-Re \& Ac-Re  & SC \\
    \midrule
    % \multirow{2}{*}{\textbf{Parameters}} & Updating facts & \multirow{2}{*}{\yes} & \multirow{2}{*}{\no} & \multirow{2}{*}{$0$} \\
    % & Training & & &\\
    \textbf{Parameters} & - & \yes & \no & 0 \\
    \midrule
    \multirow{2}{*}{\textbf{Explicit Memory}} 
     & Fixed-sized & \yes  & Partial & $O(1)$ \\
    & Flexible-sized & Partial & Partial & $o(n)$ \\
    \midrule
    \multirow{2}{*}{\textbf{Knowledge Base}}
    % \textbf{Knowledge}
    & Knowledge graph & \multirow{2}{*}{Partial} & \multirow{2}{*}{\yes} & \multirow{2}{*}{$O(n)$} \\
    % \textbf{Base} 
    & Organized Text &  & & \\
    \midrule
    \textbf{Context} & - & \no & \yes & $O(n)$ \\
    \bottomrule
\end{tabular}
\caption{In this paper, we discuss four categories of technologies with the sub-categories being sub-sections. Here ``Abs \&  ExMg''  refers to``Abstraction and Experiences Merging'', "Long-Re \& AcRe" means ``Long-Term Retention and Accurate Recalling'', and ``SC'' represents ``Storage Complexity''.}
\label{tab:structure}
% \vspace{-15pt}
\end{table*}

% \tong{Cognitive:(1) Definition; (2) The reason for including cognition to lifespan systems; (3) The solution for }

To achieve LSCS, some major challenges must be addressed: 
\textbf{(1) Abstraction and Experience Merging}: 
The system must distill experiences from its environment by extracting essential information and integrating these abstracted experiences with its existing memories. Interactions with the environment—whether through conversations~\citep{llm_annotator, dialog_summarization} or visual inputs~\citep{wang2024lvchat}—often contain redundant information that should be filtered out before storing in memory. Once the key information is abstracted, this crucial information can be used to update the system, merging new abstracted experiences with previous ones while resolving potential conflicts. 
This process enables the system to acquire new skills, deepen its understanding, correct misconceptions, and continually evolve.
\textbf{(2) Long-Term Retention and Accurate Recalling}: For systems that store past experiences in various forms—whether in memory, context, knowledge bases, or model weights—it is crucial to accurately recall relevant information in response to current queries from the environment. This capability necessitates the model’s ability to remember events from the distant past and make informed decisions based on all past experiences. 
The above two properties distinguish constructing LSCS from existing challenges such as long context problems~\citep{wang2024beyond} or continual learning~\citep{wu2024continual}, where long context problems focus on solving a problem with extremely abundant information and continual learning mainly focuses on enabling models to capture ever-changing world knowledge~\citep{zhang2023large,TemporalWiki}, facilitating self-evolution through human feedback~\citep{tao2024survey}, or adapting to specific domains~\citep{KeSLKK023, YadavSDLZTBMNRB23}.

There are existing technologies that address some of the challenges described above, but each typically falls short of the other. 
\wy{We denote the past experiences as \paex. Suppose \paex can be quantified (such as the number of tokens in the interactions between the system and the environment) and let this quantity be $n$. We then define \textbf{Storage Complexity} as a conceptual measure of the storage requirements beyond model parameters, expressed as a function of $n$, the amount of past experiences.}
These technologies can be categorized into four groups (illustrated in Figure \ref{fig:categorization}), which will be elaborated in the subsequent sections: \modelp (Section \ref{sec:save_to_model_parameters}), 
\memory (Section \ref{sec:save_to_memory}), 
\rag (Section \ref{sec:rag}), and
\lc (Section \ref{sec:save_as_raw_text}). \wy{We provide more details of the categorizations in Appendix \ref{sec:details_of_method}.}
The advantages and potential drawbacks of each approach are discussed in Section \ref{sec:discussion}, with a high-level summary provided in Table \ref{tab:structure}. This table also outlines the structure of the paper, further subdividing some categories. 

\begin{figure}[t]
    \centering
    \begin{subfigure}{0.9\textwidth}
        \centering
        \includegraphics[width=\linewidth]{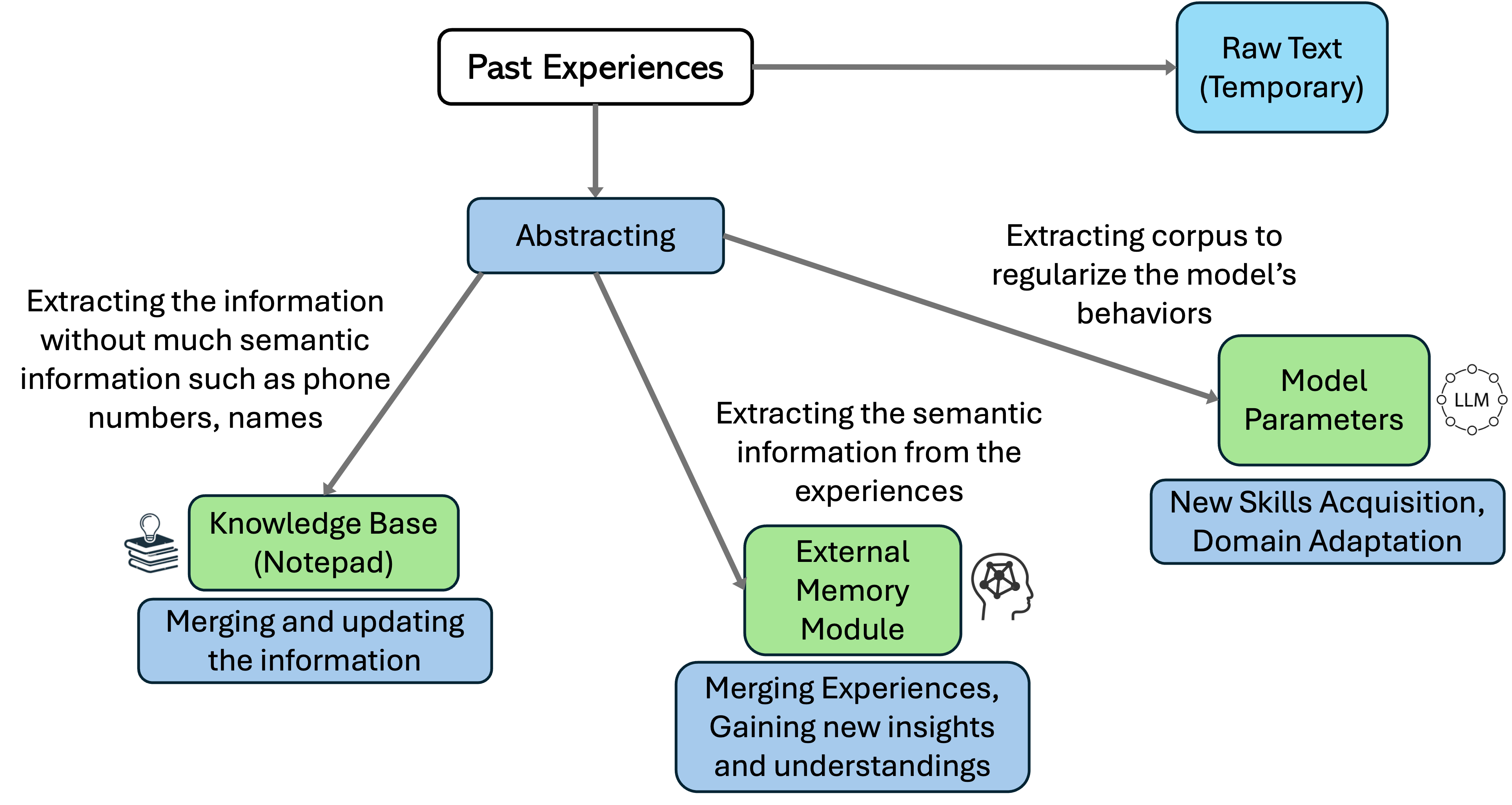}
        \caption{The process of Abstraction and Experiences Merging for LSCS.}
        \label{fig:update_process}
        % \vspace{15pt}
    \end{subfigure}
    % \hfill
    \begin{subfigure}{0.9\textwidth}
        \centering
        \includegraphics[width=\linewidth]{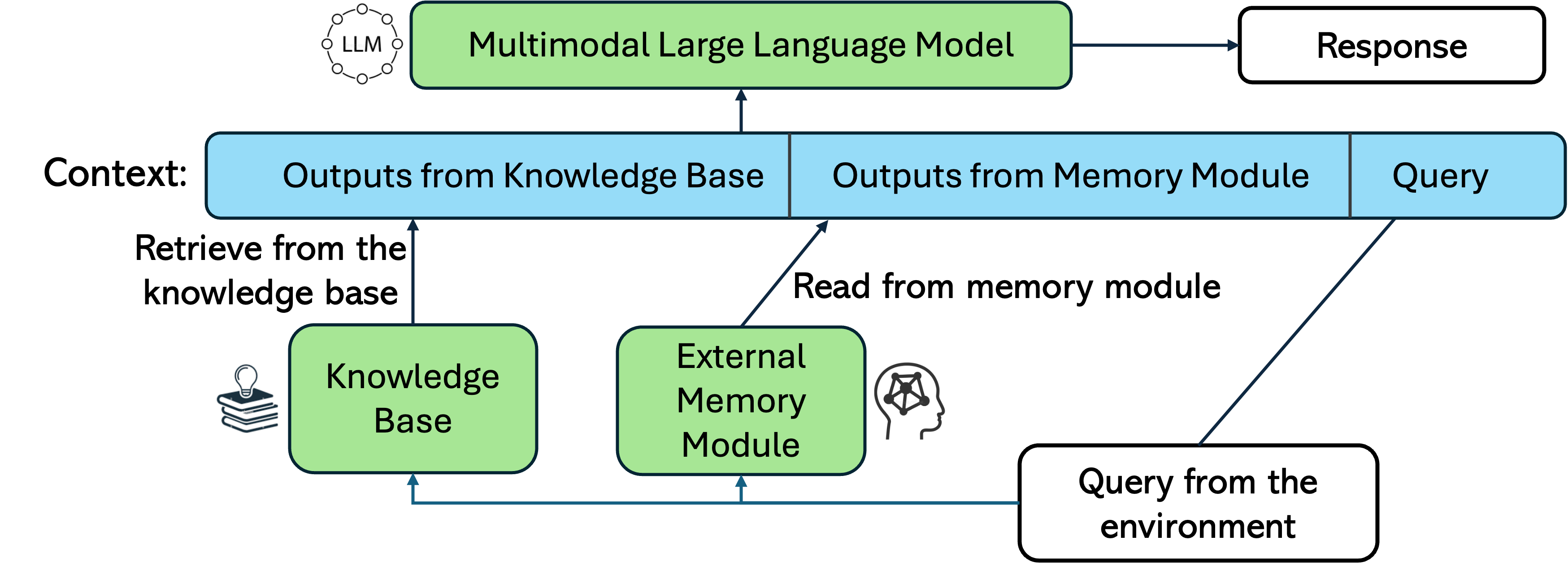}
        \caption{The process of LSCS to generate responses given the query from the environment.}
        \label{fig:generate_process}
    \end{subfigure}
    \caption{The Operating Diagram of LSCS mainly includes two parts: (a) Abstraction and Experiences Merging; (b) Generating responses according to the environmental query, where the ability of long-term retention and accurate recalling should be guaranteed. \wy{Note that we add ``Notepad'' after ``Knowledge Base'', which is meant as an analogy. See Section \ref{sub:absorbing_experiences} for more details.}}
    % \vspace{-15pt}
\label{fig:operating_mechanism_of_LSCS}
\end{figure}

With the limitations of existing technologies, we argue that any category of technologies cannot achieve LSCS alone. To deal with this issue, we propose a new possible formulation for LSCS, which integrates all the approaches above. The operational design is depicted in Figure \ref{fig:operating_mechanism_of_LSCS}, where the process is split into two stages: (1) Abstraction and Merging Experiences when new experiences arise. (2) Generating Responses based on queries from the environment, ensuring long-term retention and precise recall. Each component in this design corresponds to one of the technology categories discussed earlier, and their roles are explored in more detail in Section \ref{sec:our_suggested_instantiation_of_lscs}. 

% \vspace{-5pt}
\section{Methods of \modelp}
\label{sec:save_to_model_parameters}
% \vspace{-5pt}
The objective of integrating \paex into model parameters is allowing LLMs to continuously adapt and refine their knowledge and functionalities without repeated retraining from scratch. 
In the context of LSCS, these methods would store all the experiences in the model parameters. They do not require additional modules, thus the Storage Complexity is zero. 
The continual acquisition of experiences includes two different paths: 
model editing and continual learning, while continual learning can be further split into three main steps: continual pretraining, continual instruction tuning, and continual alignment. 
To integrate experiences into model parameters, current technologies can support injecting a large amount of domain-specific data to adapt the system via continual pre-training, or regularizing the model's behaviors via continual fine-tuning. Directly taking the various forms of inputs from the environment and storing them into model parameters are under-explored.  

% \vspace{-5pt}
\subsection{Absorbing \paex via Model Editing}
% \vspace{-5pt}
Given past experiences, it is possible to construct a knowledge graph from these experiences and then convert the triplets in the graph into factual statements, which could be further injected into the model parameters via model editing~\citep{yao2023editing}. 
Here we focus on the methods that directly make edits on the model parameters, i.e., the new knowledge is saved in parameters. Methods such as ROME~\citep{ROME}, MEMIT~\citep{memit} propose a closed-form solution to edit the MLP layers, while Model-Editing-FT~\citep{model-editing-ft} fine-tunes the whole model on the factual statements that need to be injected into the model. 
T-Patcher~\citep{TPatcher} and CaliNet~\citep{calinet} propose to store the new knowledge in additional neurons, however, \wy{storing knowledge in additional neurons may inevitably encounter the issue of ever-growing parameters in lifespan settings}.  
As model-editing methods typically focus on storing simple factual statements in the model, they are not directly applicable in the context of LSCS. Thus adjustments are needed for the application.

% \vspace{-5pt}
\subsection{Absorbing \paex via Continual Learning}
% \vspace{-5pt}
Continual learning is usually used in the following situations: (1) Performing real-time assimilation of dynamic data. Existing methods can incorporate information from various sources such as news~\citep{SunWLFTWW20,SelfInfo2023}, scholarly articles~\citep{abs-2205-09357}, and social media~\citep{abs-2205-09357}. \citet{SunWLFTWW20} present ERNIE 2.0, which is a continual pre-training framework that incrementally builds and learns from multiple tasks to maximize knowledge extraction from training data.
(2) Injecting a large amount of knowledge into the language model. \wy{Related methods using knowledge distillation to preserve existing knowledge is proposed in DER++\citet{der}.} \citet{JangYYSHKCS22} introduce continual knowledge learning, a method for updating temporal knowledge in LLMs, reducing forgetting while acquiring new information. 
(3) Domain adaptation. \citet{abs-2205-09357} continually train the model on new data streams for both language and vision, and \citet{KeSLKK023} introduce a soft-masking mechanism to update language models with domain corpora, aiming to boost performance while preserving general knowledge. Various models are proposed as the domain-adapted version of base language models such as financial domain~\citep{abs-2311-08545}, E-commerce domain~\citep{abs-2312-15696}, and academic content~\citep{abs-2311-12315}. 
These techniques could all be used for LSCS to adjust to the most up-to-date information. 
(4) Teaching the system how to speak and how to perform certain types of reasoning. To enable the models to solve novel tasks, task-incremental continual instruction tuning is proposed~\citep{abs-2403-11435-instruct,wang2023trace}. These tasks could be related to mathematical problems~\citep{azerbayev2023llemma}, calculators, search engines, and databases~\citep{qin2023toolllm}. 
With the rapid emergence of new tools like advanced software libraries, novel APIs, or domain-specific utilities ~\citep{liang2023taskmatrix,jin2023genegpt}, there is a growing need to continually update language models so they can quickly adapt and master these new tools.
(5) Adapting to evolving societal values. This leads to continual alignment, where two categories exist: (i) Updating LLMs to align to changing societal values; (ii) Incorporating new demographic groups or value systems into LLMs. Some representative works include CPPO~\citep{anonymous2023cppo}, COPF~\citep{zhang2023copf}, COPR~\citep{abs-2402-14228-copr-preference}, which aim to prevent the forgetting of old policies while injecting incremental preferences objectives into the model.
For LSCS, these techniques are important as the system may encounter situations where it needs to be up-to-date, learn plenty of new knowledge, adapt to a new domain if the system enters a new environment, learn a specific kind of tasks, or has to update to meet the societal values. 

\textbf{Catastrophic Forgetting.} One major limitation of Saving \paex into model parameters is catastrophic forgetting, which is a general problem for methods that require updating model parameters both in model-editing (e.g. in sequential editing scenarios, only LLMs with an external module can work in sequential editing tasks such as WISE~\citep{wise}, GRACE~\citep{grace} while methods without an external module typically fail~\citep{ROME,memit}) and continual learning. LLMs trained on different stages can encounter unconscious forgetting~\citep{abs-2309-06256}, eroding their general knowledge through continual instruction tuning. Studies~\citep{abs-2310-03693} have shown that the behavior of safely aligned LLMs can degrade under these conditions. Some metrics are proposed in TRACE~\citep{wang2023trace} to measure the forgetting of LLMs. 
While we want to prevent catastrophic forgetting, we do intend to have some extent of forgetting, which is aligned with the ability of \textbf{Abstraction and Experiences Merging}. Different from unlearning part of the knowledge in the LLM which is conducted after having the undesired knowledge in the model~\citep{law}, we focus on encouraging the model to abstract the information before injecting new experiences into model parameters, where it is acceptable to discard some details. However, Abstracting experiences and discarding unimportant details are not fully explored in continual learning. \wy{It is well known that a language model cannot fully remember the training data and instead extracts inherent statistical patterns from the dataset~\citep{quantifying_memorization}.} This aligns closely with our goals in LSCS, \wy{where we want to abstract events and discard unimportant details. Intuitively, it would be beneficial to control the continual learning process such that only critical information (essential for future decision-making and answering questions) is retained in the model parameters, while less relevant knowledge is discarded. Knowledge Unlearning~\citep{law} is related to this objective, as it focuses on removing specific knowledge from the model, which however, may negatively impact model performance~\citep{knowledge_unlearning_survey}. In summary, achieving controllable continual learning is beneficial but under-explored.}

% \wy{
% \subsection{Absorbing \paex via Knowledge Distillation}
% Knowledge distillation refers to the methods of designing specific training objectives (typically KL divergence) to regularize the model's behaviors in order to achieve the effects of knowledge injection. 

% }

% \vspace{-5pt}
\section{Methods of \memory}
\label{sec:save_to_memory}
% \vspace{-5pt}
Different from saving \paex into model parameters, using an external memory can help store much more experiences. The techniques in this category can be split into two sub-categories: Fixed-sized Memory Module and Flexible-sized Memory Module. As the former one has a fixed size, the storage complexity is $O(1)$. As for the latter one, because there is usually some certain form of compression and forgetting mechanisms, these works mostly have a memory pool with a size that does not grow linearly with the experiences. Thus the storage complexity for past experiences is $o(n)$. 

% \vspace{-5pt}
\subsection{Methods with a fixed-sized memory module}
% \vspace{-5pt}
\label{sub:memory_based_methods_with_a_fixed_sized_memory_pool}
Most existing works with fixed-sized memory modules have a small memory. Before the transformer era, plenty of works augment recurrent neural networks (RNNs) with a small memory module such as Memory Network~\citep{MemoryNetwork} and its follow-up \cite{E2EMN}. More advanced methods include TARDIS~\citep{gulcehre2017memory}, ARMIN~\citep{li2019armin}, Fast Weights~\citep{ba2016using}, Ordered Memory~\citep{shen2019ordered}, etc. Borrowing ideas from human memory, dual-memory system to mimic short-term and long-term memories is proposed~\citep{bidokhti2022compound,bidokhti2022dual}. Later with the introduction of transformers, Memory Transformer~\citep{memoryTransformer} adds memory tokens at the beginning of the sequence to summarize the given sequence, while RMT~\citep{RMT, bulatov2023scaling} proposes to add memory tokens both at the beginning and the end of the sequence. Here ~\cite{bulatov2023scaling} scales the number of memory tokens to 512 which enables long-term retention of previous information. 
EMMA~\citep{Memory-Enhanced-Transformer} has a memory pool of similar size, including both short-term and long-term memory. Then Larimar~\citep{larimar} proposes to use a fixed-sized episodic memory which is shown to be capable of storing up to 1000 factual statements. As opposed to small fixed-sized memory, MemoryLLM~\citep{wang2024memoryllm} introduces a memory for Llama2-7B~\citep{llama2} encapsulating 1B parameters with 7680 memory tokens per layer, enormously enlarging the memory pool. Differently, Infini-Attention~\citep{infini-attention} proposes to use linear attention to attend to the past seen tokens, where the stored matrix performs exactly like a fixed-sized memory (of size $d_{key} \times (d_{value} + 1) \times H \times L$, where $d_{key}, d_{value}, H, L$ refer to the dimension of key states, dimension of value states, number of heads and number of layers, respectively) abstracting the past tokens. 
These fixed-sized memory modules typically come with higher compression than flexible-sized memory modules and also requires much more training for deployment. 

% \vspace{-5pt}
\subsection{Methods with flexible-sized memory pool}
% \vspace{-5pt}
The flexible-sized memory pool has various forms: (1) Hidden space, where hidden states within transformers or key-value pairs are stored in the memory pool; (2) Text-based memory, where the pool consists of textual data; and (3) Symbolic space, where the memory pool contains abstracted forms such as symbols. We describe these forms of memory pools below.

\textbf{Memory Module in Hidden Space}. Some methods introduce memory slots alongside the inputs to encode information, where the number of memory slots vary according to the length of the input document~\citep{al2021memory}. Other techniques store key-value pairs in a memory pool for future retrieval such as KNN-LM~\citep{KNNLM}, Memorizing Transformer~\citep{MemoringTransformers}, LONGMEM~\citep{longMEM}, CAMELoT~\citep{he2024camelot} and Memoria~\citep{Memoria}. Here CAMELoT and Memoria include some forgetting mechanism to ensure the number of key-values pairs do not grow linearly as the input sequence becomes longer. Most recently, Memory$^3$~\citep{memory3} proposes to store the knowledge from the pretraining dataset in a knowledge base within the hidden space with $1.1\times 10^8$ text chunks with length bounded by 128 tokens. Here the number of text chunks can be easily adjusted and the number of tokens in each chunk can vary thus the size of the memory is flexible. 

\textbf{Memory Module in Text Space}. Another line of work proposes to use language models as the abstractor, which inherently extracts the information from the given input and merge them with the existing memory to form a new memory. Here the memory is in the form of a summary of the past seen document. RecurrentGPT~\citep{zhou2023recurrentgpt}, as one representative work, proposes to recurrently manage a summarization of the past context. 
Similarly, MemoryBank~\citep{MemoryBank} designs a hierarchical memory pool, storing (1) the raw conversation history between the user and the agent, (2) the summarization of the conversation, and (3) the summarization of the user's personalities. Here the last two summaries are in the form of text and serve as the compressed version of the past experiences. Different from RecurrentGPT which is specifically designed for long-context tasks, MemoryBank mainly focuses on conversations. There are also some forgetting mechanisms introduced in MemoryBank to mimic human memory and avoid the ever-growing problem of the memory pool. Moreover, SCM~\citep{wang2023enhancing} maintains the summary of each conversation and also an interaction embedding to represent the semantics of the interaction. 

\textbf{Memory Module in Symbolic Space}. The memory pool in symbolic space includes database as done in ChatDB~\citep{hu2023chatdb} where the database is updated and queried using the SQL commands generated by the fine-tuned language model. Then Voyager~\citep{voyager} maintains a code base to represent the ever-growing skill library to store and update complex behaviors in Minecraft. These strategies might have the best Long-term Retention and Accuracy Recalling ability as the past experiences are integrated explicitly in the database or the code base. However, real-world experiences encountered over a lifespan may be difficult to encode into a code base (not like the Minecraft skill set) or the database. This suggests while encoding knowledge into symbolic space could solve specific problems, its may be limited in general domains.
% \vspace{-5pt}
\section{Methods of Saving \paex into Knowledge Bases}
\label{sec:rag}
% \vspace{-5pt}
An effective strategy for LSCS involves creating knowledge bases from past experiences (\paex) that can be accessed through retrieval-augmented generation (RAG). 
In this section, we discuss two major lines of methods for storing \paex: (1) saving \paex as organized text; (2) saving \paex as knowledge graphs. 
Here saving \paex as organized text means that every detail in \paex is stored, while saving \paex as knowledge graphs involves extracting triplets from \paex and absorbing these triplets into knowledge graphs. Both approaches enable the system to maintain a repository of accumulated experiences that can be retrieved efficiently and used by LLMs during generation.

% \vspace{-10pt}
\subsection{Saving \paex as Organized Text}
% \vspace{-5pt}
% key point: every detail is saved
The first method involves saving \paex in the form of structured or organized text. This approach is similar to traditional document-based retrieval systems, where past experiences are encoded as text documents. These documents serve as retrieval units that can be accessed when needed. In LSCS, the knowledge base is continuously updated to reflect new experiences, ensuring that the LLM can access relevant information when required.
The knowledge base in this case is in the form of organized text. 

\textbf{Chunking \paex into pieces.}
Creating a text-based knowledge base involves chunking \paex into smaller, manageable units for retrieval. Common chunking strategies divide the text into fixed-sized segments based on token count~\citep{guu2020realmretrievalaugmentedlanguagemodel_realm, huang2023raven, izacard2023atlas, siriwardhana2022improvingdomainadaptationretrieval_rag_end_to_end}. Determining the optimal chunk size is crucial for effective retrieval. Large chunks (e.g., 512 tokens) can capture more context but may include irrelevant information, making it harder to pinpoint specific answers. Small chunks (e.g., 128 tokens) are more concise but may miss important contextual details necessary for accurate retrieval. Depending on the specific dataset requirements, chunking methods range from fine-grained sentence retrieval~\citep{cheng2023uprise, cheng2024lift_selfmem, jiang2023active_flare} to coarse-grained document or group of documents retrieval~\citep{shi2023replug, khattab2022demonstrate_dsp, yu2022generate_read, jiang2024longrag}. 

\textbf{Updating the text-based knowledge base.}
In LSCS, the knowledge base is updated as new experiences are encountered. This process can be done in real-time or periodically, ensuring the newest experiences are incorporated. Methods such as asynchronous re-embedding and re-indexing help keep the stored experiences up-to-date~\citep{guu2020realmretrievalaugmentedlanguagemodel_realm, huang2023raven, izacard2023atlas, siriwardhana2022improvingdomainadaptationretrieval_rag_end_to_end}. Open-source tools like LangChain and LlamaIndex facilitate real-time updates by continuously integrating new data into the vector database.

\textbf{Generating responses according to the knowledge base.} When a query is presented, the RAG mechanism retrieves relevant text from the knowledge base, which is then used to augment the LLM's prompt. This process allows the LLM to generate responses that reflect both the query and the relevant stored experiences. Text-based retrieval systems typically use either sparse keyword-based retrievers, such as BM25~\citep{robertson2009probabilistic_bm25}, or dense embedding-based retrievers, which capture semantic similarities between the query and the stored knowledge~\citep{izacard2021unsupervised_contriever, ram2021learning_spider}. Fine-tuning these retrievers can further improve their performance in domain-specific tasks~\citep{shi2023replug, zhang2023retrieve_llm_embedder, zan2022language_codegen_api, dai2022_promptagator}.

% \vspace{-10pt}
\subsection{Storing Knowledge as Knowledge Graphs} 
% \vspace{-5pt}
The second method for storing \paex is using knowledge graphs, where experiences are represented as structured data (tree or graph), allowing for a more relational organization of knowledge, capturing not just the content of experiences but also the relationships between different entities and concepts. Knowledge graphs are particularly well-suited for tasks that require reasoning over connections between pieces of information.

\textbf{Creating the knowledge graph and Retrieving}. 
To create a knowledge graph, we first need to extract entities and relations (i.e. triplets) from \paex and then transform them into graph structures, with nodes representing entities and edges representing relationships. The constructed knowledge graphs allow for richer retrieval possibilities compared to simple text-based storage. For instance, a system can retrieve not just isolated facts but also the relational context in which those facts exist, providing deeper insights during retrieval and generation tasks.
In recent works, several methods have been proposed to improve the efficiency and accuracy of graph-based retrieval~\citep{achiam2023gpt, edge2024local_graph_rag}. For instance, RAPTOR~\citep{sarthi2024raptor} and MemWalker~\citep{chen2023walking_memwalker} utilize tree-based structures to facilitate retrieval by providing contextual information at various levels of abstraction. THEANINE~\citep{kim2024theanine} and AriGraph~\citep{anokhin2024arigraphlearningknowledgegraph} construct memory graphs to organize memories for efficient retrieval. Additionally, KGP~\citep{wang2023knowledgegraphpromptingmultidocument_kgp} propose building an index across multiple documents using knowledge graphs. IIER~\citep{guo2024leveraginginterchunkinteractionsenhanced_IIER} construct a chunk-interaction graph to capture the internal connections between document chunks. Inspired by the hippocampal indexing theory of human long-term memory, 
HippoRAG~\citep{gutierrez2024hipporag} employs an LLM to process passages into knowledge graph triples and leverages the Personalized PageRank algorithm to perform context-based retrieval.  G-Retriever~\citep{he2024g_retriever} presents a retrieval approach for general textual graph tasks by formulating subgraph retrieval as a Prize-Collecting Steiner Tree 
optimization problem. 

\textbf{Updating the knowledge graph}. 
Updating knowledge graphs in LSCS involves dynamically incorporating new nodes and edges as new experiences are \wy{acquired~\citep{wang2023knowledgegraphpromptingmultidocument_kgp,guo2024leveraginginterchunkinteractionsenhanced_IIER,gutierrez2024hipporag}.} This ensures that the knowledge graph remains up-to-date and reflects the system's evolving understanding of its environment. Knowledge graphs can also be further linked with external data sources.

\textbf{Generating with the retrieved sub-graph}. 
To perform generation, 
Once the relevant sub-graph is retrieved, it is used to augment the LLM's prompt for \wy{generation~\citep{grag, subgraph_guided_knowledge}.} The structured nature of knowledge graphs provides additional benefits during this stage, as the system can leverage the explicit relationships between entities to generate more coherent and contextually informed responses. This can be particularly useful in complex reasoning tasks, where understanding the connections between concepts is as important as retrieving the correct information. 

% \vspace{-8pt}
\section{Methods of \lc}
\label{sec:save_as_raw_text}
% \vspace{-8pt}
One simple way of storing the past experiences \paex is to store all of them in the context without abstraction. In this way, the model could directly attend to these experiences when queried by the environment. As all past experiences are stored without loss of any details, the storage complexity is $O(n)$. Below we discuss the technologies that can help process the long lifespan experiences. 

% \vspace{-8pt}
\subsection{(Claimed) Length-Generalizable Architectures}
% \vspace{-5pt}

Transformers~\cite{vaswani2017attention} is the de facto mainstream backbone architecture for most modern LLMs~\cite{achiam2023gpt, touvron2023llama, llama2, gpt-j}.
To augment the self-attention layers that are position-agnostic with position information, traditional absolute positional encodings provide the absolute position information, usually with the help of a sequence of vectors called \textit{position embeddings}~\citep{vaswani2017attention, kenton2019bert, ke2020rethinking}. They, however, inherently have trouble when the model encounters unseen positions in inputs longer than the training length.
Inspired by this limitation, relative positional encoding techniques are proposed to depend on the attention logit function only on the relative distances between tokens instead of their absolute positions.
Examples include a learned attention logit bias in T5~\citep{raffel2020exploring}, Transformer-XL~\citep{dai2019Transformer}, Skyformer~\citep{skyformer2021}, Sketching~\citep{Sketching2022} and Sandwich~\citep{chi2023dissecting}, a fixed linear attention decay~\cite{press2021train}, and rotating query and key sequences based on distances such as RoPE~\citep{su2021roformer, longchat2023}, CAPE~\citep{likhomanenko2021cape} and XPos~\citep{sun2022length, ding2023longnet}. 
As a representative example widely used in multiple LLMs, RoPE \citep{su2021roformer} rotates the key and query vectors based on positions before computing the inner product. Specifically, each vector $\mathbf{x}$ (either key or query) is split into pairs of elements $\{(x_0, x_1), (x_2, x_3), \cdots\}$, with each pair interpreted as a 2-dimensional vector. 
Another example method is AliBi~\citep{press2021train}, which offsets all attention logits between tokens $i, j$ by a linear term $-m(i-j)$ and becomes $\mathbf{q}_i^\top \mathbf{k}_j - m(i-j)$. To this end, the MPT-7B codes implement an offset matrix as an additive term in attention logits. 
Both papers claimed to generalize to lengths longer than the training length. However, length generalization failures are still widely observed when directly applied to off-the-shelf Transformer-based LLMs~\citep{superhot}.
These models also suffer from overwhelming quadratic computational complexity, and the information lost issue~\citep{liu2024lost, shang2024ai}, which limits their deployment on practical computing devices and efficacy on extreme lengths.

In view of the limitations in the Transformer architecture, some papers revisit the concept of recurrent networks and structured state space models (SSMs)~\citep{guefficiently} for both length-generalizable and efficient architectures.
Recurrent structures usually have a linear computational complexity with respect to the sequence length, a desirable property for handling long contexts. They also maintain a bounded or sometimes fixed information bottleneck, which naturally prevents feature drift when length increases and can alleviate the generation quality degradation issue.
RWKV~\citep{peng2023rwkv} revisit the traditional recurrent neural networks (RNNs) and demonstrates that, contrary to the common belief, a large enough RNN can exhibit impressive performance comparable to many Transformer-based LLMs.
Similarly, Retnet~\citep{sun2023retentive} propose retentive networks, and Memba~\citep{gu2023mamba} propose to use SSMs, which combines more complicated techniques such as gating mechanism and prefix sum to improve the expressiveness and efficiency further.
However, the recurrent structure also requires a delicate tradeoff between long-term and short-term memory~\citep{gu2023mamba,xLSTM}, and their performance on long-context tasks, especially on lifespan data, which requires information retention on the order of years, is still under debate and investigation.

% \vspace{-5pt}
\subsection{Length Extension Methods for Existing LLMs}
% \vspace{-5pt}

Various methods have been proposed to extend the context length of existing LLMs, addressing their length limits and tackling the lifespan data better.
One line of work focuses on modifying the attention mechanism without changing the model parameters.
\citet{han2024lminfinite, xiao2024efficient} propose to modify the attention mechanism without changing the model parameters to enable LLMs to handle infinite context lengths. In particular, LM-Infinite~\citep{han2024lminfinite} extends LLMs to an extreme length of 200M tokens with $O(n)$ complexity and without degradation in perplexity metric while also showing improvements in downstream tasks.
\citet{jinllm} propose to group tokens into blocks and let each block share a relative position before applying the attention mechanism. They also demonstrate improved performance on long-context tasks, even though the grouping operation is not extendable to infinite size, which upper bounds the extension length.

Another line of work focuses on fine-tuning the model on longer texts to adapt to longer contexts.
As many absolute or relative position encoding methods are trained on mathematically pre-defined periodic functions, \citet{qu2023gpt, ding2024longrope, liu2023scaling, jiang2023longllmlingua} propose to apply LLMs (fine-tuned or not) on decreased period frequencies (which is equivalent to interpolating position indices) to adapt LLMs to longer sequences. This modification changes the computational features of the model, making fine-tuning necessary to adapt the model to the new context length.
Some other papers finetune LLMs with designed attention patterns~\citep{oren2024transformers, zhang2024soaring} on long contexts, using neural tangent kernel~\citep{peng2023yarn}, or with low-rank adaptation(LoRA)~\citep{chen2023longlora}.
\citet{yang2024attendre} propose a wait-to-attend mechanism to extend length limits for memory-based Transformers.
Other ways of key-value cache selection/eviction methods are investigated in \citet{ren2024efficacy, dong2024get, zhang2024h2o}.
Similarly, \citet{huang2023boosting, lee2024human} tackle long context by learning to prune, select, or summarize contexts dynamically.
Alternatively, context compression methods~\citep{shao2024flexibly} propose to learn to compress long contexts into shorter embedding sequences.
Some work proposes alternative position encodings~\citep{song2023zebra, zhang2024found, zhu2023pose} or landmark token embeddings~\citep{luo2024bge} that enable extendable context limits. 
Compared with the other methods, especially continual learning, this category of methods needs to expand the memory continuously when the system obtains new experiences, while the former could absorb these experiences into the model parameters without expanding anything.
These approaches, relying on LLMs to aggregate and process contextual information on the fly, might encounter problems in the face of mutually conflicting instructions or tasks like privacy protection and information retrieval~\citep{wang2024decodingtrust}. On the other hand, cognitive systems with an awareness of the ideal situation are expected to better maintain and handle the ongoing tasks with complete interactions.

\begin{table}[]
    \centering
    \begin{tabular}{c|cccccc}
    \toprule
    Where to store & How Compressed $\uparrow$ (1-4) & E.\texttt{Write} $\uparrow$ (1-4) & E.\texttt{Read} $\uparrow$ (1-4) \\
    \midrule
    \textbf{Parameters} & 4 & 1 & 4 \\
    \textbf{Explicit Memory}
     & 3 & 2 & 3 \\
    \textbf{Knowledge Base} & 2 & 3 & 2 \\
    \textbf{Context} & 1 & 4 & 1 \\
    \bottomrule
\end{tabular}
    \caption{Properties of different categories. ``How Compressed'' means how compressed past experiences are when stored in the corresponding form. E.\texttt{Write} and E.\texttt{Read} refer to the efficiency of \texttt{Write} process (writing the experiences into a certain form (model parameters/explicit memory/knowledge base/context)) and \texttt{Read} process (reading the experiences from a certain form), respectively. The more compressed the information is, or the higher the efficiency is, the higher score we list here.}
    \label{tab:strengths_and_weaknesses}
    % \vspace{-20pt}
\end{table}

% \vspace{-5pt}
\section{Benefits and Limitations of the Methods in Each Category}
\label{sec:discussion}
% \vspace{-5pt}

In this section, we examine the strengths and weaknesses of the various technologies mentioned above (The strengths and weaknesses are summarized in Table \ref{tab:structure} and Table \ref{tab:strengths_and_weaknesses}).

\textbf{\modelp}. 
\wy{This category includes model editing and continual learning.}
Current model editing techniques fall short in editing the model with life experiences which could be more complicated than factual statements, and continual learning methods still struggle with catastrophic forgetting. Despite these limitations, these methods still offer valuable contributions to LSCS, particularly when there is a need to inject substantial knowledge into the model, adjust its behavior, or update the system to reflect shifts in societal values.
As the model updates its parameters with new experiences, new memories are conceptually merged with existing ones. 
This process leads to automatic \textbf{Abstraction and Experiences Merging}, where the raw information is not preserved in full detail, but rather abstracted and integrated into the model. Such methods face challenges with \textbf{Long-Term Retention and Accurate Recalling} due to the inherent risk of catastrophic forgetting in continual learning and failure in sequential model editing (Although some works like WISE~\citep{wise} claim to support numerous sequential edits by using external modules, these should technically fall under the category of “\memory”). The efficiency of writing the experiences into the model parameters (E.\texttt{Write}) is low, as it involves training and back-propagation (even in model editing methods such as \cite{ROME,memit}, backpropagation is needed). In contrast, the efficiency of reading the experiences (E.\texttt{Read}) is high, as the generation speed remains unaffected by updates while the effects of the experiences can still be reflected in generation. 

\textbf{\memory}. 
Memory-based methods show promise as they allow the model to access information from the distant past by storing \paex in memory, without significantly increasing inference costs as history lengthens. These methods (1) often involve some form of abstraction, resulting in the inability to predict exact answers when queried on specific questions, as only the abstracted information from the experiences is stored. (2) may be hard to enable long-term retention as there is either explicit forgetting~\citep{wang2024memoryllm,MemoryBank} or implicit forgetting~\citep{zhou2023recurrentgpt}. One possible way to mitigate the forgetting issue is to enlarge the memory pool as studied in MemoryLLM~\citep{wang2024memoryllm}. However, the utilization of the memory pool in MemoryLLM may not be optimal, as it can only achieve around 40 steps (approximately 20k input length) of updates without completely forgetting the earlier knowledge. 
Despite the current limitations of memory-based methods, we believe memory-based methods are still an important part of achieving a powerful LSCS. In LSCS, one important ability is \textbf{Abstraction and Experiences Merging}. In memory-based methods, there are certain forms of compression when writing the experiences into the memory, 
% which is the key feature of memory-based methods.
% As mentioned above, memory-based methods usually have certain forms of compression, 
which aligns with the idea of abstraction (abstracting the incoming knowledge). 
However, experiences merging is still under-explored in memory-based methods. Though there are some preliminary explorations~\citep{he2024camelot}, how to merge the memories, especially in the hidden space is extremely hard and there are barely any works capable of merging memories in a way that similar memories are put together and even lead to new understandings instead of simply concatenating all memories. 
As for the ability of \textbf{Long-Term Retention and Accurate Recalling}, memory-based methods are inherently not able to recall accurate information as the compression and forgetting mechanisms are part of the design of the memory module, which means some detailed information is destined to be missing. However they can still achieve a long-term retention where the major contour of events could be maintained with details forgotten. 
The efficiency of the \texttt{Write} process is higher than saving \paex into model parameters as it usually requires a forward pass to convert the text into some certain space, while the efficiency of the \texttt{Read} process is lower than saving \paex into model parameters as the LLM needs to read from the memory module and put this extracted information in the context to process.

\textbf{\rag}.
storing \paex in knowledge bases is essential for achieving LSCS, enabling the system to accumulate vast amounts of knowledge over time. 
\wy{
The vast amounts of knowledge in the past can be stored as organized text or knowledge graphs, where RAG methods can be used to retrieve the relevant knowledge from the organized text or knowledge graphs~\citep{gutierrez2024hipporag,sarthi2024raptor} to augment the generation process.
}
% Whether stored as organized text or knowledge graphs, the information can be retrieved efficiently through RAG methods to inform the generation process. 
While text-based storage excels at handling large volumes of general information, knowledge graphs offer more structured and relational insights. Together, these methods allow LSCS to retrieve and utilize past experiences effectively, adapting to new situations and challenges based on accumulated knowledge.
This category includes methods that store experiences as raw text in a knowledge base or create structured representations, like graphs, by extracting triplets from the input data. When extracting triplets, 
some detailed information that cannot be represented by factual triplets is inevitably lost, and the nodes and edges in the knowledge graph can be updated to resolve conflicts and merge similar relations. However, these abstractions are fairly naive and only simple conflicts and merging can be handled. As for the knowledge base in the form of organized text, the process usually needs to abstract some knowledge from the experiences to create index such as the embedding of summary of the experiences, where the experiences merging could be even harder than using knowledge graph. 
Thus we say this category of methods involve Partial
\textbf{Abstraction and Experiences Merging}. 
These methods offer strong \textbf{Long-Term Retention and Accurate Recalling}, as the knowledge base can store extensive information. The efficiency of the \texttt{Write} process (\textbf{E.\texttt{Write}}) is higher than the above two categories, as it involves processing the input to construct structured representations which may be faster than converting all text into some other spaces. The efficiency of \texttt{Read} process (\textbf{E.\texttt{Read}}) is lower than the above two, as it requires retrieving and integrating relevant information from the knowledge base. In comparison, Saving \paex into model parameters does not require retrieval, and saving \paex into explicit memory could potentially skip the need of retrieval such as MemoryLLM~\citep{wang2024memoryllm}.

\textbf{\lc}
While saving \paex as the raw text and putting it into context is conceptually straightforward, it faces significant practical challenges.
For instance, humans, are estimated to speak an average of 16k words per day~\citep{mehl2007women}, totaling hundreds of millions of words over a lifetime~\citep{brandreth1980joy}. Consequently, current LLMs usually cannot process all past experiences. Even methods that claim to handle infinitely long contexts struggle with effectively recalling important knowledge from the past~\citep{gu2023mamba,ttt}. Despite extensive efforts to extend these length limits, obstacles such as computational inefficiency and information forgetting remain unresolved~\cite{liu2024lost}, preventing these approaches from being a fundamental solution to LSCS. 
As for the property \textbf{Abstraction and Experiences Merging} mentioned in Section \ref{sec:introduction}, this line of methods does not involve any abstraction or experiences merging. However, as all the past experiences are stored explicitly in the context, this category of methods is capable of achieving \textbf{Long-Term Retention and Accurate Recalling}. The \texttt{Write} process (\textbf{E.\texttt{Write}}) is highly efficient, as it merely involves concatenating new experiences to the existing data. However, the \texttt{Read} process (\textbf{E.\texttt{Read}}) is less efficient, as it requires the language model to process all stored experiences, leading to increased computational demands.

% \vspace{-10pt}
\section{Our Proposed Instantiation of LSCS}
\label{sec:our_suggested_instantiation_of_lscs}
% \vspace{-10pt}
Existing technologies individually address some aspects of the challenges associated with \textbf{Abstracting and Merging Experiences} and \textbf{Long-Term Retention and Accurate Recall}, but none can handle both sets of issues comprehensively. To address these limitations, we propose a new instantiation that integrates and combines these methods to create a more robust solution. As depicted in Figure \ref{fig:operating_mechanism_of_LSCS}, our instantiation operates in two key phases: absorbing experiences and generating responses to environmental queries, with the detailed descriptions shown below:

% \vspace{-8pt}
\subsection{Absorbing Experiences}
\label{sub:absorbing_experiences}
% \vspace{-8pt}
The experience absorption process within LSCS occurs with multiple levels of abstraction. This allows the system to retain both raw details and high-level summaries of experiences, which are processed and stored in several levels (corresponding to Figure \ref{fig:update_process}):

\textbf{Raw Experience Storage}: The latest experiences are captured in their raw textual form, preserving details without loss. This information is stored in the system’s memory as a direct record of events, which is accessible for immediate recall.

\textbf{Non-semantic Information Storage}: Certain types of data, such as phone numbers or some hard-to-remember names, are challenging for typical memory models to store effectively. These pieces of non-semantic information, which lack comprehensive contextual meaning but are critical for precision, are stored in a dedicated knowledge base. The knowledge base acts as a repository for factual, easily forgotten information that LSCS can access as needed. \wy{Just as humans may use notebooks to store specific pieces of information they cannot reliably and easily retain in their memory (e.g., phone numbers, exact names, addresses), an LSCS could have an external, easily accessible storage component serving a similar role. For instance, frequently updated, high-fidelity information like contact details, passcodes, or historical records (generally “verbatim” facts that are cumbersome to store in model parameters) would be stored in such a dedicated external knowledge base, thus we add the term ``Notepad'' in Figure \ref{fig:update_process} to highlight the role of the knowledge base.}

\textbf{Semantic Information Abstraction}: Experiences are also abstracted to capture their core meaning or “essence,” such as a general impression of a person, an emotional response to an event, or a rough memorization of some events that happened before. These abstracted memories, which may lack fine details, are stored in the external memory module. This module allows for memory consolidation and combination, leading to the creation of new insights and deeper understandings through a process akin to memory integration and fusion. \wy{The distinction between semantic information and non-semantic information is that non-semantic information is generally much harder to remember for humans. Humans can memorize what happened in the movie (which falls under the semantic information category) after watching it once and can recall the plot after a long time but may struggle to remember some detailed information such as phone numbers, specific names that do not have any semantic meaning (we call them non-semantic information). Thus for our instantiation, we also handle these two types of information differently.}

\textbf{Deeply Encoded Information in Model Parameters}: The highest level of abstraction occurs when repeated experiences are encoded directly into the model’s parameters, leading to the formation of long-term behaviors and habits. This mirrors how humans internalize routines or adapt to new environments over time. In LSCS, these deeply embedded patterns represent highly compressed forms of knowledge that are retained without needing further memory resources.

As experiences are abstracted through these levels, the degree of compression increases, transitioning from raw text to structured knowledge, and eventually to internalized model parameters. \wy{During this process, conflicts between new and existing knowledge may inevitably arise. Ideally, a memory model designed to resemble human memory could effectively merge similar experiences over time, while a dynamically updated knowledge graph could help resolve conflicts and integrate new information seamlessly.}

% \vspace{-8pt}
\subsection{Generating Responses to Environmental Queries}
% \vspace{-5pt}
When faced with an environmental query, LSCS gathers relevant information from all of its modules to generate a response. To begin with, the query is integrated into the context, forming the basis for the system’s reasoning. Then related information is extracted with the following processes (see Figure \ref{fig:generate_process}):

\textbf{Retrieval from Knowledge Base}: If factual or non-semantic information is required, the system queries the knowledge base for relevant data, which is then added to the context. This ensures that precise, detail-oriented information is readily available when needed.

\textbf{Abstracted Semantic Information Retrieval}: With the environmental query, the system reads the external memory module to extract the relevant semantic information which is usually in the hidden space. The extracted hidden states are also put into the context. 

\textbf{Response Generation}: With all necessary inputs in place, the large language model (potentially a multi-modal one as the experiences can include images, speech, videos as well~\citep{wang2024lvchat}). takes the context as the input and generate a response according to the context and the model parameters. This step includes reading the extracted information from the knowledge base, the memory module, and the model parameters to provide a complete, context-aware response to the environmental query.

% \vspace{-8pt}
\subsection{Summary of the Proposed Instantiation}
% \vspace{-5pt}
In the above-mentioned processes, both the knowledge base and memory module are utilized, allowing us to adopt the technologies described in \rag and \memory. Since the system incorporates a large language model, the techniques outlined in \modelp are also applicable. During response generation, the model must handle extensive contextual inputs, which can be long enough to require the large language model to have long-context processing capabilities. By leveraging these technologies, we can enable \textbf{Abstraction and Experience Merging}, where the knowledge base, external memory module, and model parameters can all store abstracted versions of experiences. Then with the presence of both the knowledge base and external memory module, the system is capable of achieving \textbf{Long-Term Retention and Accurate Recall} via querying these modules. \wy{We would like to highlight that the current instantiation of LSCS still remains conceptual and does not represent a fully implemented system. While we do want to propose a system that can actually work, there exist some challenges, illustrated in Appendix~\ref{sec:challenges_of_proposed_instantiation}. 
}
% \vspace{-8pt}
\section{Conclusion}
\label{sec:conclusion}
% \vspace{-8pt}
In this paper, we propose the concept of a \textbf{LifeSpan Cognitive System} (LSCS), which aims to manage rapid, incremental learning while retaining the ability to recall previous experiences. 
To achieve LSCS, we argue there are two major challenges: (1) Abstraction and Experiences Merging, (2) Long-Term Retention and Accurate Recalling. Existing technologies have the potential to partially solve these challenges. 
Based on our defined metric, \textbf{Storage Complexity}, for saving past experiences (\paex), we categorize existing technologies into four distinct classes. We summarize various technologies within each category and discuss the strengths and weaknesses of the methods in each category. We argue that achieving an LSCS requires a nuanced approach integrating multiple strategies to address the above two challenges. To this end, we propose a conceptual instantiation for LSCS. We hope to provide insights into LSCS and encourage further efforts.
% to build more human-like cognitive systems. 

\section*{Acknowledgement}
We thank the reviewers and the action editor for their valuable suggestions and comments. This research is based upon work supported by U.S. DARPA ECOLE Program No. HR00112390060. The views and conclusions contained herein are those of the authors and should not be interpreted as necessarily representing the official policies, either expressed or implied, of DARPA, or the U.S. Government. The U.S. Government is authorized to reproduce and distribute reprints for governmental purposes notwithstanding any copyright annotation therein.

\bibliography{main}

\begin{thebibliography}{151}
\providecommand{\natexlab}[1]{#1}
\providecommand{\url}[1]{\texttt{#1}}
\expandafter\ifx\csname urlstyle\endcsname\relax
  \providecommand{\doi}[1]{doi: #1}\else
  \providecommand{\doi}{doi: \begingroup \urlstyle{rm}\Url}\fi

\bibitem[Achiam et~al.(2023)Achiam, Adler, Agarwal, Ahmad, Akkaya, Aleman, Almeida, Altenschmidt, Altman, Anadkat, et~al.]{achiam2023gpt}
Josh Achiam, Steven Adler, Sandhini Agarwal, Lama Ahmad, Ilge Akkaya, Florencia~Leoni Aleman, Diogo Almeida, Janko Altenschmidt, Sam Altman, Shyamal Anadkat, et~al.
\newblock Gpt-4 technical report.
\newblock \emph{arXiv preprint arXiv:2303.08774}, 2023.

\bibitem[Al~Adel \& Burtsev(2021)Al~Adel and Burtsev]{al2021memory}
Arij Al~Adel and Mikhail~S Burtsev.
\newblock Memory transformer with hierarchical attention for long document processing.
\newblock In \emph{2021 International Conference Engineering and Telecommunication (En\&T)}, pp.\  1--7. IEEE, 2021.

\bibitem[Anokhin et~al.(2024)Anokhin, Semenov, Sorokin, Evseev, Burtsev, and Burnaev]{anokhin2024arigraphlearningknowledgegraph}
Petr Anokhin, Nikita Semenov, Artyom Sorokin, Dmitry Evseev, Mikhail Burtsev, and Evgeny Burnaev.
\newblock Arigraph: Learning knowledge graph world models with episodic memory for llm agents, 2024.
\newblock URL \url{https://arxiv.org/abs/2407.04363}.

\bibitem[Azerbayev et~al.(2023)Azerbayev, Schoelkopf, Paster, et~al.]{azerbayev2023llemma}
Zhangir Azerbayev, Hailey Schoelkopf, Keiran Paster, et~al.
\newblock Llemma: An open language model for mathematics.
\newblock \emph{CoRR}, 2023.

\bibitem[Ba et~al.(2016)Ba, Hinton, Mnih, Leibo, and Ionescu]{ba2016using}
Jimmy Ba, Geoffrey~E Hinton, Volodymyr Mnih, Joel~Z Leibo, and Catalin Ionescu.
\newblock Using fast weights to attend to the recent past.
\newblock \emph{Advances in neural information processing systems}, 29, 2016.

\bibitem[Beck et~al.(2024)Beck, P{\"{o}}ppel, Spanring, Auer, Prudnikova, Kopp, Klambauer, Brandstetter, and Hochreiter]{xLSTM}
Maximilian Beck, Korbinian P{\"{o}}ppel, Markus Spanring, Andreas Auer, Oleksandra Prudnikova, Michael Kopp, G{\"{u}}nter Klambauer, Johannes Brandstetter, and Sepp Hochreiter.
\newblock xlstm: Extended long short-term memory.
\newblock \emph{CoRR}, abs/2405.04517, 2024.

\bibitem[Bidokhti \& Ghaemmaghami(2022{\natexlab{a}})Bidokhti and Ghaemmaghami]{bidokhti2022compound}
Amir Bidokhti and Shahrokh Ghaemmaghami.
\newblock Compound short-and long-term memory for memory augmented neural networks.
\newblock \emph{Engineering Applications of Artificial Intelligence}, 116:\penalty0 105450, 2022{\natexlab{a}}.

\bibitem[Bidokhti \& Ghaemmaghami(2022{\natexlab{b}})Bidokhti and Ghaemmaghami]{bidokhti2022dual}
Amir Bidokhti and Shahrokh Ghaemmaghami.
\newblock Dual memory structure for memory augmented neural networks for question-answering tasks.
\newblock In \emph{2022 12th International Conference on Computer and Knowledge Engineering (ICCKE)}, pp.\  142--147. IEEE, 2022{\natexlab{b}}.

\bibitem[Brandreth(1980)]{brandreth1980joy}
Gyles~Daubeney Brandreth.
\newblock The joy of lex: How to have fun with 860,341,500 words.
\newblock \emph{(No Title)}, 1980.

\bibitem[Bulatov et~al.(2022)Bulatov, Kuratov, and Burtsev]{RMT}
Aydar Bulatov, Yuri Kuratov, and Mikhail~S. Burtsev.
\newblock Recurrent memory transformer.
\newblock In \emph{NeurIPS}, 2022.

\bibitem[Bulatov et~al.(2023)Bulatov, Kuratov, and Burtsev]{bulatov2023scaling}
Aydar Bulatov, Yuri Kuratov, and Mikhail~S Burtsev.
\newblock Scaling transformer to 1m tokens and beyond with rmt.
\newblock \emph{arXiv preprint arXiv:2304.11062}, 2023.

\bibitem[Burtsev \& Sapunov(2020)Burtsev and Sapunov]{memoryTransformer}
Mikhail~S. Burtsev and Grigory~V. Sapunov.
\newblock Memory transformer.
\newblock \emph{CoRR}, abs/2006.11527, 2020.
\newblock URL \url{https://arxiv.org/abs/2006.11527}.

\bibitem[Buzzega et~al.(2020)Buzzega, Boschini, Porrello, Abati, and Calderara]{der}
Pietro Buzzega, Matteo Boschini, Angelo Porrello, Davide Abati, and Simone Calderara.
\newblock Dark experience for general continual learning: a strong, simple baseline.
\newblock In \emph{NeurIPS}, 2020.

\bibitem[Carlini et~al.(2023)Carlini, Ippolito, Jagielski, Lee, Tram{\`{e}}r, and Zhang]{quantifying_memorization}
Nicholas Carlini, Daphne Ippolito, Matthew Jagielski, Katherine Lee, Florian Tram{\`{e}}r, and Chiyuan Zhang.
\newblock Quantifying memorization across neural language models.
\newblock In \emph{{ICLR}}. OpenReview.net, 2023.

\bibitem[Chen et~al.(2023{\natexlab{a}})Chen, Pasunuru, Weston, and Celikyilmaz]{chen2023walking_memwalker}
Howard Chen, Ramakanth Pasunuru, Jason Weston, and Asli Celikyilmaz.
\newblock Walking down the memory maze: Beyond context limit through interactive reading.
\newblock \emph{arXiv preprint arXiv:2310.05029}, 2023{\natexlab{a}}.

\bibitem[Chen et~al.(2021)Chen, Zeng, Ji, and Yang]{skyformer2021}
Yifan Chen, Qi~Zeng, Heng Ji, and Yun Yang.
\newblock Skyformer: Remodel self-attention with gaussian kernel and nyström method.
\newblock In \emph{Proc. Thirty-fifth Annual Conference on Neural Information Processing Systems (NeurIPS2021)}, 2021.

\bibitem[Chen et~al.(2022)Chen, Zeng, Ji, and Yang]{Sketching2022}
Yifan Chen, Qi~Zeng, Heng Ji, and Yun Yang.
\newblock Sketching as a tool for understanding and accelerating self-attention for long sequences.
\newblock In \emph{Proc. The 2022 Conference of the North American Chapter of the Association for Computational Linguistics - Human Language Technologies (NAACL-HLT2022)}, 2022.

\bibitem[Chen et~al.(2024)Chen, Wu, and Zaki]{subgraph_guided_knowledge}
Yu~Chen, Lingfei Wu, and Mohammed~J. Zaki.
\newblock Toward subgraph-guided knowledge graph question generation with graph neural networks.
\newblock \emph{{IEEE} Trans. Neural Networks Learn. Syst.}, 35\penalty0 (9):\penalty0 12706--12717, 2024.

\bibitem[Chen et~al.(2023{\natexlab{b}})Chen, Qian, Tang, Lai, Liu, Han, and Jia]{chen2023longlora}
Yukang Chen, Shengju Qian, Haotian Tang, Xin Lai, Zhijian Liu, Song Han, and Jiaya Jia.
\newblock Longlora: Efficient fine-tuning of long-context large language models.
\newblock In \emph{The Twelfth International Conference on Learning Representations}, 2023{\natexlab{b}}.

\bibitem[Cheng et~al.(2023)Cheng, Huang, Bi, Zhan, Liu, Wang, Sun, Wei, Deng, and Zhang]{cheng2023uprise}
Daixuan Cheng, Shaohan Huang, Junyu Bi, Yuefeng Zhan, Jianfeng Liu, Yujing Wang, Hao Sun, Furu Wei, Denvy Deng, and Qi~Zhang.
\newblock Uprise: Universal prompt retrieval for improving zero-shot evaluation.
\newblock \emph{arXiv preprint arXiv:2303.08518}, 2023.

\bibitem[Cheng et~al.(2024)Cheng, Luo, Chen, Liu, Zhao, and Yan]{cheng2024lift_selfmem}
Xin Cheng, Di~Luo, Xiuying Chen, Lemao Liu, Dongyan Zhao, and Rui Yan.
\newblock Lift yourself up: Retrieval-augmented text generation with self-memory.
\newblock \emph{Advances in Neural Information Processing Systems}, 36, 2024.

\bibitem[Chi et~al.(2023)Chi, Fan, Rudnicky, and Ramadge]{chi2023dissecting}
Ta-Chung Chi, Ting-Han Fan, Alexander Rudnicky, and Peter Ramadge.
\newblock Dissecting transformer length extrapolation via the lens of receptive field analysis.
\newblock In \emph{Proceedings of the 61st Annual Meeting of the Association for Computational Linguistics (Volume 1: Long Papers)}, pp.\  13522--13537, 2023.

\bibitem[Cossu et~al.(2022)Cossu, Tuytelaars, Carta, et~al.]{abs-2205-09357}
Andrea Cossu, Tinne Tuytelaars, Antonio Carta, et~al.
\newblock Continual pre-training mitigates forgetting in language and vision.
\newblock \emph{CoRR}, 2022.

\bibitem[Dai et~al.(2022)Dai, Zhao, Ma, Luan, Ni, Lu, Bakalov, Guu, Hall, and Chang]{dai2022_promptagator}
Zhuyun Dai, Vincent~Y Zhao, Ji~Ma, Yi~Luan, Jianmo Ni, Jing Lu, Anton Bakalov, Kelvin Guu, Keith~B Hall, and Ming-Wei Chang.
\newblock Promptagator: Few-shot dense retrieval from 8 examples.
\newblock \emph{arXiv preprint arXiv:2209.11755}, 2022.

\bibitem[Dai et~al.(2019)Dai, Yang, Yang, Carbonell, Le, and Salakhutdinov]{dai2019Transformer}
Zihang Dai, Zhilin Yang, Yiming Yang, Jaime~G Carbonell, Quoc Le, and Ruslan Salakhutdinov.
\newblock Transformer-xl: Attentive language models beyond a fixed-length context.
\newblock In \emph{Proceedings of the 57th Annual Meeting of the Association for Computational Linguistics}, pp.\  2978--2988, 2019.

\bibitem[Das et~al.(2024)Das, Chaudhury, Nelson, Melnyk, Swaminathan, Dai, Lozano, Kollias, Chenthamarakshan, Navr{\'{a}}til, Dan, and Chen]{larimar}
Payel Das, Subhajit Chaudhury, Elliot Nelson, Igor Melnyk, Sarathkrishna Swaminathan, Sihui Dai, Aur{\'{e}}lie~C. Lozano, Georgios Kollias, Vijil Chenthamarakshan, Jir{\'{\i}} Navr{\'{a}}til, Soham Dan, and Pin{-}Yu Chen.
\newblock Larimar: Large language models with episodic memory control.
\newblock In \emph{{ICML}}. OpenReview.net, 2024.

\bibitem[Ding et~al.(2023)Ding, Ma, Dong, Zhang, Huang, Wang, and Wei]{ding2023longnet}
Jiayu Ding, Shuming Ma, Li~Dong, Xingxing Zhang, Shaohan Huang, Wenhui Wang, and Furu Wei.
\newblock Longnet: Scaling transformers to 1,000,000,000 tokens.
\newblock \emph{arXiv preprint arXiv:2307.02486}, 2023.

\bibitem[Ding et~al.(2024)Ding, Zhang, Zhang, Xu, Shang, Xu, Yang, and Yang]{ding2024longrope}
Yiran Ding, Li~Lyna Zhang, Chengruidong Zhang, Yuanyuan Xu, Ning Shang, Jiahang Xu, Fan Yang, and Mao Yang.
\newblock Longrope: Extending llm context window beyond 2 million tokens.
\newblock \emph{arXiv preprint arXiv:2402.13753}, 2024.

\bibitem[Dong et~al.(2024)Dong, Yang, Zhang, Wang, Chi, and Chen]{dong2024get}
Harry Dong, Xinyu Yang, Zhenyu Zhang, Zhangyang Wang, Yuejie Chi, and Beidi Chen.
\newblock Get more with less: Synthesizing recurrence with kv cache compression for efficient llm inference.
\newblock \emph{arXiv preprint arXiv:2402.09398}, 2024.

\bibitem[Dong et~al.(2022)Dong, Dai, Song, Xu, Sui, and Li]{calinet}
Qingxiu Dong, Damai Dai, Yifan Song, Jingjing Xu, Zhifang Sui, and Lei Li.
\newblock Calibrating factual knowledge in pretrained language models.
\newblock In \emph{{EMNLP} (Findings)}, pp.\  5937--5947. Association for Computational Linguistics, 2022.

\bibitem[Edge et~al.(2024)Edge, Trinh, Cheng, Bradley, Chao, Mody, Truitt, and Larson]{edge2024local_graph_rag}
Darren Edge, Ha~Trinh, Newman Cheng, Joshua Bradley, Alex Chao, Apurva Mody, Steven Truitt, and Jonathan Larson.
\newblock From local to global: A graph rag approach to query-focused summarization.
\newblock \emph{arXiv preprint arXiv:2404.16130}, 2024.

\bibitem[Feng et~al.(2021)Feng, Feng, Qin, Qin, and Liu]{llm_annotator}
Xiachong Feng, Xiaocheng Feng, Libo Qin, Bing Qin, and Ting Liu.
\newblock Language model as an annotator: Exploring dialogpt for dialogue summarization.
\newblock In \emph{{ACL/IJCNLP} {(1)}}, pp.\  1479--1491. Association for Computational Linguistics, 2021.

\bibitem[Feng et~al.(2022)Feng, Feng, and Qin]{dialog_summarization}
Xiachong Feng, Xiaocheng Feng, and Bing Qin.
\newblock A survey on dialogue summarization: Recent advances and new frontiers.
\newblock In \emph{{IJCAI}}, pp.\  5453--5460. ijcai.org, 2022.

\bibitem[Gangadhar \& Stratos(2024)Gangadhar and Stratos]{model-editing-ft}
Govind Gangadhar and Karl Stratos.
\newblock Model editing by pure fine-tuning.
\newblock \emph{CoRR}, abs/2402.11078, 2024.

\bibitem[Gu \& Dao(2023)Gu and Dao]{gu2023mamba}
Albert Gu and Tri Dao.
\newblock Mamba: Linear-time sequence modeling with selective state spaces.
\newblock \emph{arXiv preprint arXiv:2312.00752}, 2023.

\bibitem[Gu et~al.()Gu, Goel, and Re]{guefficiently}
Albert Gu, Karan Goel, and Christopher Re.
\newblock Efficiently modeling long sequences with structured state spaces.
\newblock In \emph{International Conference on Learning Representations}.

\bibitem[Gulcehre et~al.(2017)Gulcehre, Chandar, and Bengio]{gulcehre2017memory}
Caglar Gulcehre, Sarath Chandar, and Yoshua Bengio.
\newblock Memory augmented neural networks with wormhole connections.
\newblock \emph{arXiv preprint arXiv:1701.08718}, 2017.

\bibitem[Guo et~al.(2024)Guo, Wang, Liu, Tang, Li, Xu, Yang, Gao, Li, and Wen]{guo2024leveraginginterchunkinteractionsenhanced_IIER}
Tiezheng Guo, Chen Wang, Yanyi Liu, Jiawei Tang, Pan Li, Sai Xu, Qingwen Yang, Xianlin Gao, Zhi Li, and Yingyou Wen.
\newblock Leveraging inter-chunk interactions for enhanced retrieval in large language model-based question answering, 2024.
\newblock URL \url{https://arxiv.org/abs/2408.02907}.

\bibitem[Guti{\'e}rrez et~al.(2024)Guti{\'e}rrez, Shu, Gu, Yasunaga, and Su]{gutierrez2024hipporag}
Bernal~Jim{\'e}nez Guti{\'e}rrez, Yiheng Shu, Yu~Gu, Michihiro Yasunaga, and Yu~Su.
\newblock Hipporag: Neurobiologically inspired long-term memory for large language models.
\newblock \emph{arXiv preprint arXiv:2405.14831}, 2024.

\bibitem[Guu et~al.(2020)Guu, Lee, Tung, Pasupat, and Chang]{guu2020realmretrievalaugmentedlanguagemodel_realm}
Kelvin Guu, Kenton Lee, Zora Tung, Panupong Pasupat, and Ming-Wei Chang.
\newblock Realm: Retrieval-augmented language model pre-training, 2020.
\newblock URL \url{https://arxiv.org/abs/2002.08909}.

\bibitem[Han et~al.(2024)Han, Wang, Peng, Xiong, Chen, Ji, and Wang]{han2024lminfinite}
Chi Han, Qifan Wang, Hao Peng, Wenhan Xiong, Yu~Chen, Heng Ji, and Sinong Wang.
\newblock Lm-infinite: Zero-shot extreme length generalization for large language models.
\newblock In \emph{Proceedings of the 14th International Joint Conference on Natural Language Processing}, pp.\  23--28, 2024.

\bibitem[Hartvigsen et~al.(2022)Hartvigsen, Sankaranarayanan, Palangi, Kim, and Ghassemi]{grace}
Thomas Hartvigsen, Swami Sankaranarayanan, Hamid Palangi, Yoon Kim, and Marzyeh Ghassemi.
\newblock Aging with grace: Lifelong model editing with discrete key-value adaptors.
\newblock \emph{arXiv preprint arXiv:2211.11031}, 2022.

\bibitem[He et~al.(2024{\natexlab{a}})He, Tian, Sun, Chawla, Laurent, LeCun, Bresson, and Hooi]{he2024g_retriever}
Xiaoxin He, Yijun Tian, Yifei Sun, Nitesh~V Chawla, Thomas Laurent, Yann LeCun, Xavier Bresson, and Bryan Hooi.
\newblock G-retriever: Retrieval-augmented generation for textual graph understanding and question answering.
\newblock \emph{arXiv preprint arXiv:2402.07630}, 2024{\natexlab{a}}.

\bibitem[He et~al.(2024{\natexlab{b}})He, Karlinsky, Kim, McAuley, Krotov, and Feris]{he2024camelot}
Zexue He, Leonid Karlinsky, Donghyun Kim, Julian McAuley, Dmitry Krotov, and Rogerio Feris.
\newblock Camelot: Towards large language models with training-free consolidated associative memory.
\newblock \emph{arXiv preprint arXiv:2402.13449}, 2024{\natexlab{b}}.

\bibitem[Hu et~al.(2023)Hu, Fu, Du, Luo, Zhao, and Zhao]{hu2023chatdb}
Chenxu Hu, Jie Fu, Chenzhuang Du, Simian Luo, Junbo Zhao, and Hang Zhao.
\newblock Chatdb: Augmenting llms with databases as their symbolic memory.
\newblock \emph{arXiv preprint arXiv:2306.03901}, 2023.

\bibitem[Hu et~al.(2024)Hu, Lei, Zhang, Pan, Ling, and Zhao]{grag}
Yuntong Hu, Zhihan Lei, Zheng Zhang, Bo~Pan, Chen Ling, and Liang Zhao.
\newblock {GRAG:} graph retrieval-augmented generation.
\newblock \emph{CoRR}, abs/2405.16506, 2024.

\bibitem[Huang et~al.(2023{\natexlab{a}})Huang, Ping, Xu, Shoeybi, Chang, and Catanzaro]{huang2023raven}
Jie Huang, Wei Ping, Peng Xu, Mohammad Shoeybi, Kevin Chen-Chuan Chang, and Bryan Catanzaro.
\newblock Raven: In-context learning with retrieval augmented encoder-decoder language models.
\newblock \emph{arXiv preprint arXiv:2308.07922}, 2023{\natexlab{a}}.

\bibitem[Huang et~al.(2023{\natexlab{b}})Huang, Zhang, Cheng, and Yang]{huang2023boosting}
Xijie Huang, Li~Lyna Zhang, Kwang-Ting Cheng, and Mao Yang.
\newblock Boosting llm reasoning: Push the limits of few-shot learning with reinforced in-context pruning.
\newblock \emph{arXiv preprint arXiv:2312.08901}, 2023{\natexlab{b}}.

\bibitem[Huang et~al.(2023{\natexlab{c}})Huang, Shen, Zhang, Zhou, Rong, and Xiong]{TPatcher}
Zeyu Huang, Yikang Shen, Xiaofeng Zhang, Jie Zhou, Wenge Rong, and Zhang Xiong.
\newblock Transformer-patcher: One mistake worth one neuron.
\newblock \emph{arXiv preprint arXiv:2301.09785}, 2023{\natexlab{c}}.

\bibitem[Izacard et~al.(2021)Izacard, Caron, Hosseini, Riedel, Bojanowski, Joulin, and Grave]{izacard2021unsupervised_contriever}
Gautier Izacard, Mathilde Caron, Lucas Hosseini, Sebastian Riedel, Piotr Bojanowski, Armand Joulin, and Edouard Grave.
\newblock Unsupervised dense information retrieval with contrastive learning.
\newblock \emph{arXiv preprint arXiv:2112.09118}, 2021.

\bibitem[Izacard et~al.(2023)Izacard, Lewis, Lomeli, Hosseini, Petroni, Schick, Dwivedi-Yu, Joulin, Riedel, and Grave]{izacard2023atlas}
Gautier Izacard, Patrick Lewis, Maria Lomeli, Lucas Hosseini, Fabio Petroni, Timo Schick, Jane Dwivedi-Yu, Armand Joulin, Sebastian Riedel, and Edouard Grave.
\newblock Atlas: Few-shot learning with retrieval augmented language models.
\newblock \emph{Journal of Machine Learning Research}, 24\penalty0 (251):\penalty0 1--43, 2023.

\bibitem[Jang et~al.(2022{\natexlab{a}})Jang, Ye, Lee, Yang, Shin, Han, Kim, and Seo]{TemporalWiki}
Joel Jang, Seonghyeon Ye, Changho Lee, Sohee Yang, Joongbo Shin, Janghoon Han, Gyeonghun Kim, and Minjoon Seo.
\newblock Temporalwiki: {A} lifelong benchmark for training and evaluating ever-evolving language models.
\newblock In \emph{{EMNLP}}, pp.\  6237--6250. Association for Computational Linguistics, 2022{\natexlab{a}}.

\bibitem[Jang et~al.(2022{\natexlab{b}})Jang, Ye, Yang, et~al.]{JangYYSHKCS22}
Joel Jang, Seonghyeon Ye, Sohee Yang, et~al.
\newblock Towards continual knowledge learning of language models.
\newblock In \emph{ICLR}, 2022{\natexlab{b}}.

\bibitem[Jiang et~al.(2023{\natexlab{a}})Jiang, Wu, Luo, Li, Lin, Yang, and Qiu]{jiang2023longllmlingua}
Huiqiang Jiang, Qianhui Wu, Xufang Luo, Dongsheng Li, Chin-Yew Lin, Yuqing Yang, and Lili Qiu.
\newblock Longllmlingua: Accelerating and enhancing llms in long context scenarios via prompt compression.
\newblock \emph{arXiv preprint arXiv:2310.06839}, 2023{\natexlab{a}}.

\bibitem[Jiang et~al.(2023{\natexlab{b}})Jiang, Xu, Gao, Sun, Liu, Dwivedi-Yu, Yang, Callan, and Neubig]{jiang2023active_flare}
Zhengbao Jiang, Frank~F Xu, Luyu Gao, Zhiqing Sun, Qian Liu, Jane Dwivedi-Yu, Yiming Yang, Jamie Callan, and Graham Neubig.
\newblock Active retrieval augmented generation.
\newblock \emph{arXiv preprint arXiv:2305.06983}, 2023{\natexlab{b}}.

\bibitem[Jiang et~al.(2024)Jiang, Ma, and Chen]{jiang2024longrag}
Ziyan Jiang, Xueguang Ma, and Wenhu Chen.
\newblock Longrag: Enhancing retrieval-augmented generation with long-context llms.
\newblock \emph{arXiv preprint arXiv:2406.15319}, 2024.

\bibitem[Jin et~al.(2024)Jin, Han, Yang, Jiang, Liu, Chang, Chen, and Hu]{jinllm}
Hongye Jin, Xiaotian Han, Jingfeng Yang, Zhimeng Jiang, Zirui Liu, Chia{-}Yuan Chang, Huiyuan Chen, and Xia Hu.
\newblock {LLM} maybe longlm: Self-extend {LLM} context window without tuning.
\newblock \emph{CoRR}, abs/2401.01325, 2024.

\bibitem[Jin et~al.(2023)Jin, Yang, Chen, and Lu]{jin2023genegpt}
Qiao Jin, Yifan Yang, Qingyu Chen, and Zhiyong Lu.
\newblock Genegpt: Augmenting large language models with domain tools for improved access to biomedical information.
\newblock \emph{arXiv:2304.09667}, 2023.

\bibitem[Kaiokendev(2023)]{superhot}
Kaiokendev.
\newblock Things i\'m learning while training superhot.
\newblock \url{https://kaiokendev.github.io/til#extending-context-to-8k}, 2023.

\bibitem[Ke et~al.(2020)Ke, He, and Liu]{ke2020rethinking}
Guolin Ke, Di~He, and Tie-Yan Liu.
\newblock Rethinking positional encoding in language pre-training.
\newblock In \emph{International Conference on Learning Representations}, 2020.

\bibitem[Ke et~al.(2023)Ke, Shao, Lin, Konishi, Kim, and Liu]{KeSLKK023}
Zixuan Ke, Yijia Shao, Haowei Lin, Tatsuya Konishi, Gyuhak Kim, and Bing Liu.
\newblock Continual pre-training of language models.
\newblock In \emph{ICLR}, 2023.

\bibitem[Kenton \& Toutanova(2019)Kenton and Toutanova]{kenton2019bert}
Jacob Devlin Ming-Wei~Chang Kenton and Lee~Kristina Toutanova.
\newblock Bert: Pre-training of deep bidirectional transformers for language understanding.
\newblock In \emph{Proceedings of NAACL-HLT}, pp.\  4171--4186, 2019.

\bibitem[Khandelwal et~al.(2019)Khandelwal, Levy, Jurafsky, Zettlemoyer, and Lewis]{KNNLM}
Urvashi Khandelwal, Omer Levy, Dan Jurafsky, Luke Zettlemoyer, and Mike Lewis.
\newblock Generalization through memorization: Nearest neighbor language models.
\newblock \emph{arXiv preprint arXiv:1911.00172}, 2019.

\bibitem[Khattab et~al.(2022)Khattab, Santhanam, Li, Hall, Liang, Potts, and Zaharia]{khattab2022demonstrate_dsp}
Omar Khattab, Keshav Santhanam, Xiang~Lisa Li, David Hall, Percy Liang, Christopher Potts, and Matei Zaharia.
\newblock Demonstrate-search-predict: Composing retrieval and language models for knowledge-intensive nlp.
\newblock \emph{arXiv preprint arXiv:2212.14024}, 2022.

\bibitem[Kim et~al.(2024)Kim, Ong, Kwon, Kim, Ka, Bae, Jo, Hwang, Lee, and Yeo]{kim2024theanine}
Seo~Hyun Kim, Kai Tzu-iunn Ong, Taeyoon Kwon, Namyoung Kim, Keummin Ka, SeongHyeon Bae, Yohan Jo, Seung-won Hwang, Dongha Lee, and Jinyoung Yeo.
\newblock Theanine: Revisiting memory management in long-term conversations with timeline-augmented response generation.
\newblock \emph{arXiv preprint arXiv:2406.10996}, 2024.

\bibitem[Lee et~al.(2024)Lee, Chen, Furuta, Canny, and Fischer]{lee2024human}
Kuang-Huei Lee, Xinyun Chen, Hiroki Furuta, John Canny, and Ian Fischer.
\newblock A human-inspired reading agent with gist memory of very long contexts.
\newblock \emph{arXiv preprint arXiv:2402.09727}, 2024.

\bibitem[Li et~al.(2023)Li, Shao, Xie, Sheng, Zheng, Gonzalez, Stoica, Ma, and Zhang]{longchat2023}
Dacheng Li, Rulin Shao, Anze Xie, Ying Sheng, Lianmin Zheng, Joseph~E. Gonzalez, Ion Stoica, Xuezhe Ma, and Hao Zhang.
\newblock How long can open-source llms truly promise on context length?, June 2023.
\newblock URL \url{https://lmsys.org/blog/2023-06-29-longchat}.

\bibitem[Li et~al.(2019)Li, Zhong, Huang, Zhang, Li, and Li]{li2019armin}
Zhangheng Li, Jia-Xing Zhong, Jingjia Huang, Tao Zhang, Thomas Li, and Ge~Li.
\newblock Armin: Towards a more efficient and light-weight recurrent memory network.
\newblock \emph{arXiv preprint arXiv:1906.12087}, 2019.

\bibitem[Liang et~al.(2023)Liang, Wu, Song, et~al.]{liang2023taskmatrix}
Yaobo Liang, Chenfei Wu, Ting Song, et~al.
\newblock Taskmatrix. ai: Completing tasks by connecting foundation models with millions of apis.
\newblock \emph{arXiv:2303.16434}, 2023.

\bibitem[Likhomanenko et~al.(2021)Likhomanenko, Xu, Synnaeve, Collobert, and Rogozhnikov]{likhomanenko2021cape}
Tatiana Likhomanenko, Qiantong Xu, Gabriel Synnaeve, Ronan Collobert, and Alex Rogozhnikov.
\newblock Cape: Encoding relative positions with continuous augmented positional embeddings.
\newblock \emph{Advances in Neural Information Processing Systems}, 34:\penalty0 16079--16092, 2021.

\bibitem[Lin et~al.(2023)Lin, Tan, Lin, et~al.]{abs-2309-06256}
Yong Lin, Lu~Tan, Hangyu Lin, et~al.
\newblock Speciality vs generality: An empirical study on catastrophic forgetting in fine-tuning foundation models.
\newblock \emph{arXiv:2309.06256}, 2023.

\bibitem[Liu et~al.(2024)Liu, Lin, Hewitt, Paranjape, Bevilacqua, Petroni, and Liang]{liu2024lost}
Nelson~F Liu, Kevin Lin, John Hewitt, Ashwin Paranjape, Michele Bevilacqua, Fabio Petroni, and Percy Liang.
\newblock Lost in the middle: How language models use long contexts.
\newblock \emph{Transactions of the Association for Computational Linguistics}, 12, 2024.

\bibitem[Liu et~al.(2023)Liu, Yan, An, Qiu, and Lin]{liu2023scaling}
Xiaoran Liu, Hang Yan, Chenxin An, Xipeng Qiu, and Dahua Lin.
\newblock Scaling laws of rope-based extrapolation.
\newblock In \emph{The Twelfth International Conference on Learning Representations}, 2023.

\bibitem[Luo et~al.(2024)Luo, Liu, Xiao, and Liu]{luo2024bge}
Kun Luo, Zheng Liu, Shitao Xiao, and Kang Liu.
\newblock Bge landmark embedding: A chunking-free embedding method for retrieval augmented long-context large language models.
\newblock \emph{arXiv preprint arXiv:2402.11573}, 2024.

\bibitem[Ma et~al.(2023)Ma, Huang, Huang, et~al.]{abs-2312-15696}
Shirong Ma, Shen Huang, Shulin Huang, et~al.
\newblock Ecomgpt-ct: Continual pre-training of e-commerce large language models with semi-structured data.
\newblock \emph{CoRR}, 2023.

\bibitem[Mehl et~al.(2007)Mehl, Vazire, Ram{\'\i}rez-Esparza, Slatcher, and Pennebaker]{mehl2007women}
Matthias~R Mehl, Simine Vazire, Nair{\'a}n Ram{\'\i}rez-Esparza, Richard~B Slatcher, and James~W Pennebaker.
\newblock Are women really more talkative than men?
\newblock \emph{Science}, 317\penalty0 (5834):\penalty0 82--82, 2007.

\bibitem[Meng et~al.(2022)Meng, Bau, Andonian, and Belinkov]{ROME}
Kevin Meng, David Bau, Alex Andonian, and Yonatan Belinkov.
\newblock Locating and editing factual associations in gpt.
\newblock \emph{Advances in Neural Information Processing Systems}, 35:\penalty0 17359--17372, 2022.

\bibitem[Meng et~al.(2023)Meng, Sharma, Andonian, Belinkov, and Bau]{memit}
Kevin Meng, Arnab~Sen Sharma, Alex~J. Andonian, Yonatan Belinkov, and David Bau.
\newblock Mass-editing memory in a transformer.
\newblock In \emph{{ICLR}}. OpenReview.net, 2023.

\bibitem[Moro et~al.(2023)Moro, Ragazzi, Valgimigli, Frisoni, Sartori, and Marfia]{Memory-Enhanced-Transformer}
Gianluca Moro, Luca Ragazzi, Lorenzo Valgimigli, Giacomo Frisoni, Claudio Sartori, and Gustavo Marfia.
\newblock Efficient memory-enhanced transformer for long-document summarization in low-resource regimes.
\newblock \emph{Sensors}, 23\penalty0 (7):\penalty0 3542, 2023.

\bibitem[Munkhdalai et~al.(2024)Munkhdalai, Faruqui, and Gopal]{infini-attention}
Tsendsuren Munkhdalai, Manaal Faruqui, and Siddharth Gopal.
\newblock Leave no context behind: Efficient infinite context transformers with infini-attention.
\newblock \emph{CoRR}, abs/2404.07143, 2024.

\bibitem[Oren et~al.(2024)Oren, Hassid, Adi, and Schwartz]{oren2024transformers}
Matanel Oren, Michael Hassid, Yossi Adi, and Roy Schwartz.
\newblock Transformers are multi-state rnns.
\newblock \emph{arXiv preprint arXiv:2401.06104}, 2024.

\bibitem[Park et~al.(2023)Park, O'Brien, Cai, Morris, Liang, and Bernstein]{park2023generative}
Joon~Sung Park, Joseph O'Brien, Carrie~Jun Cai, Meredith~Ringel Morris, Percy Liang, and Michael~S Bernstein.
\newblock Generative agents: Interactive simulacra of human behavior.
\newblock In \emph{Proceedings of the 36th annual acm symposium on user interface software and technology}, pp.\  1--22, 2023.

\bibitem[Park \& Bak(2024)Park and Bak]{Memoria}
Sangjun Park and JinYeong Bak.
\newblock Memoria: Resolving fateful forgetting problem through human-inspired memory architecture, 2024.

\bibitem[Peng et~al.()Peng, Alcaide, Anthony, Albalak, Arcadinho, Biderman, Cao, Cheng, Chung, Derczynski, et~al.]{peng2023rwkv}
Bo~Peng, Eric Alcaide, Quentin~Gregory Anthony, Alon Albalak, Samuel Arcadinho, Stella Biderman, Huanqi Cao, Xin Cheng, Michael~Nguyen Chung, Leon Derczynski, et~al.
\newblock Rwkv: Reinventing rnns for the transformer era.
\newblock In \emph{The 2023 Conference on Empirical Methods in Natural Language Processing}.

\bibitem[Peng et~al.(2023)Peng, Quesnelle, Fan, and Shippole]{peng2023yarn}
Bowen Peng, Jeffrey Quesnelle, Honglu Fan, and Enrico Shippole.
\newblock Yarn: Efficient context window extension of large language models.
\newblock In \emph{The Twelfth International Conference on Learning Representations}, 2023.

\bibitem[Press et~al.(2021)Press, Smith, and Lewis]{press2021train}
Ofir Press, Noah Smith, and Mike Lewis.
\newblock Train short, test long: Attention with linear biases enables input length extrapolation.
\newblock In \emph{International Conference on Learning Representations}, 2021.

\bibitem[Qi et~al.(2023)Qi, Zeng, Xie, et~al.]{abs-2310-03693}
Xiangyu Qi, Yi~Zeng, Tinghao Xie, et~al.
\newblock Fine-tuning aligned language models compromises safety, even when users do not intend to!
\newblock \emph{arXiv:2310.03693}, 2023.

\bibitem[Qin et~al.(2023)Qin, Liang, Ye, et~al.]{qin2023toolllm}
Yujia Qin, Shihao Liang, Yining Ye, et~al.
\newblock Toolllm: Facilitating large language models to master 16000+ real-world apis.
\newblock \emph{arXiv:2307.16789}, 2023.

\bibitem[Qu(2023)]{qu2023gpt}
Zhijie Qu.
\newblock Gpt rotational position embedding for length extrapolation.
\newblock In \emph{Proceedings of the 2023 6th International Conference on Machine Learning and Natural Language Processing}, pp.\  166--170, 2023.

\bibitem[Raffel et~al.(2020)Raffel, Shazeer, Roberts, Lee, Narang, Matena, Zhou, Li, and Liu]{raffel2020exploring}
Colin Raffel, Noam Shazeer, Adam Roberts, Katherine Lee, Sharan Narang, Michael Matena, Yanqi Zhou, Wei Li, and Peter~J Liu.
\newblock Exploring the limits of transfer learning with a unified text-to-text transformer.
\newblock \emph{The Journal of Machine Learning Research}, 21\penalty0 (1):\penalty0 5485--5551, 2020.

\bibitem[Ram et~al.(2021)Ram, Shachaf, Levy, Berant, and Globerson]{ram2021learning_spider}
Ori Ram, Gal Shachaf, Omer Levy, Jonathan Berant, and Amir Globerson.
\newblock Learning to retrieve passages without supervision.
\newblock \emph{arXiv preprint arXiv:2112.07708}, 2021.

\bibitem[Ren \& Zhu(2024)Ren and Zhu]{ren2024efficacy}
Siyu Ren and Kenny~Q Zhu.
\newblock On the efficacy of eviction policy for key-value constrained generative language model inference.
\newblock \emph{arXiv preprint arXiv:2402.06262}, 2024.

\bibitem[Robertson et~al.(2009)Robertson, Zaragoza, et~al.]{robertson2009probabilistic_bm25}
Stephen Robertson, Hugo Zaragoza, et~al.
\newblock The probabilistic relevance framework: Bm25 and beyond.
\newblock \emph{Foundations and Trends{\textregistered} in Information Retrieval}, 3\penalty0 (4):\penalty0 333--389, 2009.

\bibitem[Sarthi et~al.(2024)Sarthi, Abdullah, Tuli, Khanna, Goldie, and Manning]{sarthi2024raptor}
Parth Sarthi, Salman Abdullah, Aditi Tuli, Shubh Khanna, Anna Goldie, and Christopher~D Manning.
\newblock Raptor: Recursive abstractive processing for tree-organized retrieval.
\newblock \emph{arXiv preprint arXiv:2401.18059}, 2024.

\bibitem[Shang et~al.(2024)Shang, Zheng, Ying, Tao, and Team]{shang2024ai}
Jingbo Shang, Zai Zheng, Xiang Ying, Felix Tao, and Mindverse Team.
\newblock Ai-native memory: A pathway from llms towards agi.
\newblock \emph{arXiv preprint arXiv:2406.18312}, 2024.

\bibitem[Shao et~al.(2024)Shao, Xiao, Liu, and Zhang]{shao2024flexibly}
Ninglu Shao, Shitao Xiao, Zheng Liu, and Peitian Zhang.
\newblock Flexibly scaling large language models contexts through extensible tokenization.
\newblock \emph{arXiv preprint arXiv:2401.07793}, 2024.

\bibitem[Shen et~al.(2019)Shen, Tan, Hosseini, Lin, Sordoni, and Courville]{shen2019ordered}
Yikang Shen, Shawn Tan, Arian Hosseini, Zhouhan Lin, Alessandro Sordoni, and Aaron~C Courville.
\newblock Ordered memory.
\newblock \emph{Advances in Neural Information Processing Systems}, 32, 2019.

\bibitem[Shi et~al.(2023)Shi, Min, Yasunaga, Seo, James, Lewis, Zettlemoyer, and Yih]{shi2023replug}
Weijia Shi, Sewon Min, Michihiro Yasunaga, Minjoon Seo, Rich James, Mike Lewis, Luke Zettlemoyer, and Wen-tau Yih.
\newblock Replug: Retrieval-augmented black-box language models.
\newblock \emph{arXiv preprint arXiv:2301.12652}, 2023.

\bibitem[Si et~al.(2023)Si, Zhang, Chang, Zhang, Qu, and Zhang]{knowledge_unlearning_survey}
Nianwen Si, Hao Zhang, Heyu Chang, Wenlin Zhang, Dan Qu, and Weiqiang Zhang.
\newblock Knowledge unlearning for llms: Tasks, methods, and challenges.
\newblock \emph{CoRR}, abs/2311.15766, 2023.

\bibitem[Siriwardhana et~al.(2022)Siriwardhana, Weerasekera, Wen, Kaluarachchi, Rana, and Nanayakkara]{siriwardhana2022improvingdomainadaptationretrieval_rag_end_to_end}
Shamane Siriwardhana, Rivindu Weerasekera, Elliott Wen, Tharindu Kaluarachchi, Rajib Rana, and Suranga Nanayakkara.
\newblock Improving the domain adaptation of retrieval augmented generation (rag) models for open domain question answering, 2022.
\newblock URL \url{https://arxiv.org/abs/2210.02627}.

\bibitem[Song et~al.(2023)Song, Wang, Cho, Pan, and Yu]{song2023zebra}
Kaiqiang Song, Xiaoyang Wang, Sangwoo Cho, Xiaoman Pan, and Dong Yu.
\newblock Zebra: Extending context window with layerwise grouped local-global attention.
\newblock \emph{arXiv preprint arXiv:2312.08618}, 2023.

\bibitem[Su et~al.(2021)Su, Lu, Pan, Murtadha, Wen, and Liu]{su2021roformer}
Jianlin Su, Yu~Lu, Shengfeng Pan, Ahmed Murtadha, Bo~Wen, and Yunfeng Liu.
\newblock Roformer: Enhanced transformer with rotary position embedding.
\newblock \emph{arXiv preprint arXiv:2104.09864}, 2021.

\bibitem[Sukhbaatar et~al.(2015)Sukhbaatar, Weston, Fergus, et~al.]{E2EMN}
Sainbayar Sukhbaatar, Jason Weston, Rob Fergus, et~al.
\newblock End-to-end memory networks.
\newblock \emph{Advances in neural information processing systems}, 28, 2015.

\bibitem[Sun et~al.(2020)Sun, Wang, Li, et~al.]{SunWLFTWW20}
Yu~Sun, Shuohuan Wang, Yu{-}Kun Li, et~al.
\newblock {ERNIE} 2.0: {A} continual pre-training framework for language understanding.
\newblock In \emph{AAAI}, 2020.

\bibitem[Sun et~al.(2024)Sun, Li, Dalal, Xu, Vikram, Zhang, Dubois, Chen, Wang, Koyejo, et~al.]{ttt}
Yu~Sun, Xinhao Li, Karan Dalal, Jiarui Xu, Arjun Vikram, Genghan Zhang, Yann Dubois, Xinlei Chen, Xiaolong Wang, Sanmi Koyejo, et~al.
\newblock Learning to (learn at test time): Rnns with expressive hidden states.
\newblock \emph{arXiv preprint arXiv:2407.04620}, 2024.

\bibitem[Sun et~al.(2022)Sun, Dong, Patra, Ma, Huang, Benhaim, Chaudhary, Song, and Wei]{sun2022length}
Yutao Sun, Li~Dong, Barun Patra, Shuming Ma, Shaohan Huang, Alon Benhaim, Vishrav Chaudhary, Xia Song, and Furu Wei.
\newblock A length-extrapolatable transformer.
\newblock \emph{arXiv preprint arXiv:2212.10554}, 2022.

\bibitem[Sun et~al.(2023)Sun, Dong, Huang, Ma, Xia, Xue, Wang, and Wei]{sun2023retentive}
Yutao Sun, Li~Dong, Shaohan Huang, Shuming Ma, Yuqing Xia, Jilong Xue, Jianyong Wang, and Furu Wei.
\newblock Retentive network: A successor to transformer for large language models.
\newblock \emph{arXiv preprint arXiv:2307.08621}, 2023.

\bibitem[Tao et~al.(2024)Tao, Lin, Chen, Li, Wu, Li, Jin, Huang, Tao, and Zhou]{tao2024survey}
Zhengwei Tao, Ting-En Lin, Xiancai Chen, Hangyu Li, Yuchuan Wu, Yongbin Li, Zhi Jin, Fei Huang, Dacheng Tao, and Jingren Zhou.
\newblock A survey on self-evolution of large language models.
\newblock \emph{arXiv preprint arXiv:2404.14387}, 2024.

\bibitem[Touvron et~al.(2023{\natexlab{a}})Touvron, Lavril, Izacard, Martinet, Lachaux, Lacroix, Rozi{\`e}re, Goyal, Hambro, Azhar, et~al.]{touvron2023llama}
Hugo Touvron, Thibaut Lavril, Gautier Izacard, Xavier Martinet, Marie-Anne Lachaux, Timoth{\'e}e Lacroix, Baptiste Rozi{\`e}re, Naman Goyal, Eric Hambro, Faisal Azhar, et~al.
\newblock Llama: Open and efficient foundation language models.
\newblock \emph{arXiv preprint arXiv:2302.13971}, 2023{\natexlab{a}}.

\bibitem[Touvron et~al.(2023{\natexlab{b}})Touvron, Martin, Stone, Albert, Almahairi, Babaei, Bashlykov, Batra, Bhargava, Bhosale, et~al.]{llama2}
Hugo Touvron, Louis Martin, Kevin Stone, Peter Albert, Amjad Almahairi, Yasmine Babaei, Nikolay Bashlykov, Soumya Batra, Prajjwal Bhargava, Shruti Bhosale, et~al.
\newblock Llama 2: Open foundation and fine-tuned chat models.
\newblock \emph{arXiv preprint arXiv:2307.09288}, 2023{\natexlab{b}}.

\bibitem[Vaswani et~al.(2017)Vaswani, Shazeer, Parmar, Uszkoreit, Jones, Gomez, Kaiser, and Polosukhin]{vaswani2017attention}
Ashish Vaswani, Noam Shazeer, Niki Parmar, Jakob Uszkoreit, Llion Jones, Aidan~N Gomez, {\L}ukasz Kaiser, and Illia Polosukhin.
\newblock Attention is all you need.
\newblock \emph{Advances in neural information processing systems}, 30, 2017.

\bibitem[Wang \& Komatsuzaki(2021)Wang and Komatsuzaki]{gpt-j}
Ben Wang and Aran Komatsuzaki.
\newblock {GPT-J-6B: A 6 Billion Parameter Autoregressive Language Model}.
\newblock \url{https://github.com/kingoflolz/mesh-transformer-jax}, May 2021.

\bibitem[Wang et~al.(2023{\natexlab{a}})Wang, Liang, Yang, Huang, Wu, Wu, Lu, Ma, and Li]{wang2023enhancing}
Bing Wang, Xinnian Liang, Jian Yang, Hui Huang, Shuangzhi Wu, Peihao Wu, Lu~Lu, Zejun Ma, and Zhoujun Li.
\newblock Enhancing large language model with self-controlled memory framework.
\newblock \emph{arXiv e-prints}, pp.\  arXiv--2304, 2023{\natexlab{a}}.

\bibitem[Wang et~al.(2024{\natexlab{a}})Wang, Chen, Pei, Xie, Kang, Zhang, Xu, Xiong, Dutta, Schaeffer, et~al.]{wang2024decodingtrust}
Boxin Wang, Weixin Chen, Hengzhi Pei, Chulin Xie, Mintong Kang, Chenhui Zhang, Chejian Xu, Zidi Xiong, Ritik Dutta, Rylan Schaeffer, et~al.
\newblock Decodingtrust: A comprehensive assessment of trustworthiness in gpt models.
\newblock \emph{Advances in Neural Information Processing Systems}, 36, 2024{\natexlab{a}}.

\bibitem[Wang et~al.(2023{\natexlab{b}})Wang, Xie, Jiang, Mandlekar, Xiao, Zhu, Fan, and Anandkumar]{voyager}
Guanzhi Wang, Yuqi Xie, Yunfan Jiang, Ajay Mandlekar, Chaowei Xiao, Yuke Zhu, Linxi Fan, and Anima Anandkumar.
\newblock Voyager: An open-ended embodied agent with large language models, 2023{\natexlab{b}}.

\bibitem[Wang et~al.(2024{\natexlab{b}})Wang, Chen, Cheng, Liao, Zhang, Wu, Yu, Xu, Zhang, Luo, Li, Yang, Huang, and Li]{loong}
Minzheng Wang, Longze Chen, Fu~Cheng, Shengyi Liao, Xinghua Zhang, Bingli Wu, Haiyang Yu, Nan Xu, Lei Zhang, Run Luo, Yunshui Li, Min Yang, Fei Huang, and Yongbin Li.
\newblock Leave no document behind: Benchmarking long-context llms with extended multi-doc {QA}.
\newblock In \emph{{EMNLP}}, pp.\  5627--5646. Association for Computational Linguistics, 2024{\natexlab{b}}.

\bibitem[Wang et~al.(2024{\natexlab{c}})Wang, Li, Zhang, Xu, Yao, Jiang, Xie, Huang, and Chen]{wise}
Peng Wang, Zexi Li, Ningyu Zhang, Ziwen Xu, Yunzhi Yao, Yong Jiang, Pengjun Xie, Fei Huang, and Huajun Chen.
\newblock {WISE:} rethinking the knowledge memory for lifelong model editing of large language models.
\newblock \emph{CoRR}, abs/2405.14768, 2024{\natexlab{c}}.

\bibitem[Wang et~al.(2023{\natexlab{c}})Wang, Dong, Cheng, Liu, Yan, Gao, and Wei]{longMEM}
Weizhi Wang, Li~Dong, Hao Cheng, Xiaodong Liu, Xifeng Yan, Jianfeng Gao, and Furu Wei.
\newblock Augmenting language models with long-term memory.
\newblock \emph{arXiv preprint arXiv:2306.07174}, 2023{\natexlab{c}}.

\bibitem[Wang et~al.(2023{\natexlab{d}})Wang, Zhang, Chen, et~al.]{wang2023trace}
Xiao Wang, Yuansen Zhang, Tianze Chen, et~al.
\newblock Trace: A comprehensive benchmark for continual learning in large language models.
\newblock \emph{CoRR}, 2023{\natexlab{d}}.

\bibitem[Wang et~al.(2024{\natexlab{d}})Wang, Salmani, Omidi, Ren, Rezagholizadeh, and Eshaghi]{wang2024beyond}
Xindi Wang, Mahsa Salmani, Parsa Omidi, Xiangyu Ren, Mehdi Rezagholizadeh, and Armaghan Eshaghi.
\newblock Beyond the limits: A survey of techniques to extend the context length in large language models.
\newblock \emph{arXiv preprint arXiv:2402.02244}, 2024{\natexlab{d}}.

\bibitem[Wang et~al.(2024{\natexlab{e}})Wang, Liu, Shi, Li, Chen, Lu, and Yang]{abs-2403-11435-instruct}
Yifan Wang, Yafei Liu, Chufan Shi, Haoling Li, Chen Chen, Haonan Lu, and Yujiu Yang.
\newblock Inscl: {A} data-efficient continual learning paradigm for fine-tuning large language models with instructions.
\newblock \emph{CoRR}, abs/2403.11435, 2024{\natexlab{e}}.
\newblock URL \url{https://doi.org/10.48550/arXiv.2403.11435}.

\bibitem[Wang et~al.(2023{\natexlab{e}})Wang, Lipka, Rossi, Siu, Zhang, and Derr]{wang2023knowledgegraphpromptingmultidocument_kgp}
Yu~Wang, Nedim Lipka, Ryan~A. Rossi, Alexa Siu, Ruiyi Zhang, and Tyler Derr.
\newblock Knowledge graph prompting for multi-document question answering, 2023{\natexlab{e}}.
\newblock URL \url{https://arxiv.org/abs/2308.11730}.

\bibitem[Wang et~al.(2024{\natexlab{f}})Wang, Chen, Shang, and McAuley]{wang2024memoryllm}
Yu~Wang, Xiusi Chen, Jingbo Shang, and Julian McAuley.
\newblock Memoryllm: Towards self-updatable large language models.
\newblock \emph{arXiv preprint arXiv:2402.04624}, 2024{\natexlab{f}}.

\bibitem[Wang et~al.(2024{\natexlab{g}})Wang, Wu, He, Chen, and McAuley]{law}
Yu~Wang, Ruihan Wu, Zexue He, Xiusi Chen, and Julian McAuley.
\newblock Large scale knowledge washing.
\newblock \emph{arXiv preprint arXiv:2405.16720}, 2024{\natexlab{g}}.

\bibitem[Wang et~al.(2024{\natexlab{h}})Wang, Zhang, McAuley, and He]{wang2024lvchat}
Yu~Wang, Zeyuan Zhang, Julian McAuley, and Zexue He.
\newblock Lvchat: Facilitating long video comprehension.
\newblock \emph{arXiv preprint arXiv:2402.12079}, 2024{\natexlab{h}}.

\bibitem[Wei et~al.(2023)Wei, Xu, Qi, et~al.]{abs-2311-12315}
Shufa Wei, Xiaolong Xu, Xianbiao Qi, et~al.
\newblock Academicgpt: Empowering academic research.
\newblock \emph{CoRR}, 2023.

\bibitem[Weston et~al.(2014)Weston, Chopra, and Bordes]{MemoryNetwork}
Jason Weston, Sumit Chopra, and Antoine Bordes.
\newblock Memory networks.
\newblock \emph{arXiv preprint arXiv:1410.3916}, 2014.

\bibitem[Wu et~al.(2024)Wu, Luo, Li, Pan, Vu, and Haffari]{wu2024continual}
Tongtong Wu, Linhao Luo, Yuan-Fang Li, Shirui Pan, Thuy-Trang Vu, and Gholamreza Haffari.
\newblock Continual learning for large language models: A survey.
\newblock \emph{arXiv preprint arXiv:2402.01364}, 2024.

\bibitem[Wu et~al.(2022)Wu, Rabe, Hutchins, and Szegedy]{MemoringTransformers}
Yuhuai Wu, Markus~Norman Rabe, DeLesley Hutchins, and Christian Szegedy.
\newblock Memorizing transformers.
\newblock In \emph{The Tenth International Conference on Learning Representations, {ICLR} 2022, Virtual Event, April 25-29, 2022}. OpenReview.net, 2022.
\newblock URL \url{https://openreview.net/forum?id=TrjbxzRcnf-}.

\bibitem[Xiao et~al.(2024)Xiao, Tian, Chen, Han, and Lewis]{xiao2024efficient}
Guangxuan Xiao, Yuandong Tian, Beidi Chen, Song Han, and Mike Lewis.
\newblock Efficient streaming language models with attention sinks.
\newblock In \emph{The Twelfth International Conference on Learning Representations}, 2024.
\newblock URL \url{https://openreview.net/forum?id=NG7sS51zVF}.

\bibitem[Xie et~al.(2023)Xie, Aggarwal, and Ahmad]{abs-2311-08545}
Yong Xie, Karan Aggarwal, and Aitzaz Ahmad.
\newblock Efficient continual pre-training for building domain specific large language models.
\newblock \emph{CoRR}, 2023.

\bibitem[Yadav et~al.(2023)Yadav, Sun, Ding, et~al.]{YadavSDLZTBMNRB23}
Prateek Yadav, Qing Sun, Hantian Ding, et~al.
\newblock Exploring continual learning for code generation models.
\newblock In \emph{ACL}, 2023.

\bibitem[Yang et~al.(2024)Yang, Lin, Wang, Wu, Li, Tang, Wei, Wang, Tang, Song, Xi, Yu, Chen, Xiong, Tang, and E]{memory3}
Hongkang Yang, Zehao Lin, Wenjin Wang, Hao Wu, Zhiyu Li, Bo~Tang, Wenqiang Wei, Jinbo Wang, Zeyun Tang, Shichao Song, Chenyang Xi, Yu~Yu, Kai Chen, Feiyu Xiong, Linpeng Tang, and Weinan E.
\newblock Memory\({}^{\mbox{3}}\): Language modeling with explicit memory.
\newblock \emph{CoRR}, abs/2407.01178, 2024.

\bibitem[Yang \& Hua(2024)Yang and Hua]{yang2024attendre}
Zi~Yang and Nan Hua.
\newblock Attendre: Wait to attend by retrieval with evicted queries in memory-based transformers for long context processing.
\newblock \emph{arXiv preprint arXiv:2401.04881}, 2024.

\bibitem[Yao et~al.(2023)Yao, Wang, Tian, Cheng, Li, Deng, Chen, and Zhang]{yao2023editing}
Yunzhi Yao, Peng Wang, Bozhong Tian, Siyuan Cheng, Zhoubo Li, Shumin Deng, Huajun Chen, and Ningyu Zhang.
\newblock Editing large language models: Problems, methods, and opportunities.
\newblock \emph{arXiv preprint arXiv:2305.13172}, 2023.

\bibitem[Yu \& Ji(2023)Yu and Ji]{SelfInfo2023}
Pengfei Yu and Heng Ji.
\newblock Self information update for large language models through mitigating exposure bias.
\newblock In \emph{arxiv}, 2023.

\bibitem[Yu et~al.(2022)Yu, Iter, Wang, Xu, Ju, Sanyal, Zhu, Zeng, and Jiang]{yu2022generate_read}
Wenhao Yu, Dan Iter, Shuohang Wang, Yichong Xu, Mingxuan Ju, Soumya Sanyal, Chenguang Zhu, Michael Zeng, and Meng Jiang.
\newblock Generate rather than retrieve: Large language models are strong context generators.
\newblock \emph{arXiv preprint arXiv:2209.10063}, 2022.

\bibitem[Zan et~al.(2022)Zan, Chen, Lin, Guan, Wang, and Lou]{zan2022language_codegen_api}
Daoguang Zan, Bei Chen, Zeqi Lin, Bei Guan, Yongji Wang, and Jian-Guang Lou.
\newblock When language model meets private library.
\newblock \emph{arXiv preprint arXiv:2210.17236}, 2022.

\bibitem[Zhang et~al.(2023{\natexlab{a}})Zhang, Gui, Zhai, et~al.]{zhang2023copf}
Han Zhang, Lin Gui, Yuanzhao Zhai, et~al.
\newblock Copf: Continual learning human preference through optimal policy fitting.
\newblock \emph{arXiv:2310.15694}, 2023{\natexlab{a}}.

\bibitem[Zhang et~al.(2024{\natexlab{a}})Zhang, Gui, Lei, Zhai, Zhang, He, Wang, Yu, Wong, Liang, and Xu]{abs-2402-14228-copr-preference}
Han Zhang, Lin Gui, Yu~Lei, Yuanzhao Zhai, Yehong Zhang, Yulan He, Hui Wang, Yue Yu, Kam{-}Fai Wong, Bin Liang, and Ruifeng Xu.
\newblock {COPR:} continual human preference learning via optimal policy regularization.
\newblock \emph{CoRR}, abs/2402.14228, 2024{\natexlab{a}}.
\newblock URL \url{https://doi.org/10.48550/arXiv.2402.14228}.

\bibitem[Zhang et~al.(2024{\natexlab{b}})Zhang, Lei, Gui, Yang, He, Wang, and Xu]{anonymous2023cppo}
Han Zhang, Yu~Lei, Lin Gui, Min Yang, Yulan He, Hui Wang, and Ruifeng Xu.
\newblock {CPPO}: Continual learning for reinforcement learning with human feedback.
\newblock In \emph{The Twelfth International Conference on Learning Representations}, 2024{\natexlab{b}}.
\newblock URL \url{https://openreview.net/forum?id=86zAUE80pP}.

\bibitem[Zhang et~al.(2023{\natexlab{b}})Zhang, Xiao, Liu, Dou, and Nie]{zhang2023retrieve_llm_embedder}
Peitian Zhang, Shitao Xiao, Zheng Liu, Zhicheng Dou, and Jian-Yun Nie.
\newblock Retrieve anything to augment large language models.
\newblock \emph{arXiv preprint arXiv:2310.07554}, 2023{\natexlab{b}}.

\bibitem[Zhang et~al.(2024{\natexlab{c}})Zhang, Liu, Xiao, Shao, Ye, and Dou]{zhang2024soaring}
Peitian Zhang, Zheng Liu, Shitao Xiao, Ninglu Shao, Qiwei Ye, and Zhicheng Dou.
\newblock Soaring from 4k to 400k: Extending llm's context with activation beacon.
\newblock \emph{arXiv preprint arXiv:2401.03462}, 2024{\natexlab{c}}.

\bibitem[Zhang et~al.(2024{\natexlab{d}})Zhang, Chen, Hu, Xu, Chen, Hao, Han, Thai, Wang, Liu, et~al.]{zhang2024infty}
Xinrong Zhang, Yingfa Chen, Shengding Hu, Zihang Xu, Junhao Chen, Moo~Khai Hao, Xu~Han, Zhen~Leng Thai, Shuo Wang, Zhiyuan Liu, et~al.
\newblock $\infty$ bench: Extending long context evaluation beyond 100k tokens.
\newblock \emph{arXiv preprint arXiv:2402.13718}, 2024{\natexlab{d}}.

\bibitem[Zhang et~al.(2024{\natexlab{e}})Zhang, Chen, Liu, Yao, Ruwase, Chen, Wu, and Wang]{zhang2024found}
Zhenyu Zhang, Runjin Chen, Shiwei Liu, Zhewei Yao, Olatunji Ruwase, Beidi Chen, Xiaoxia Wu, and Zhangyang Wang.
\newblock Found in the middle: How language models use long contexts better via plug-and-play positional encoding.
\newblock \emph{arXiv preprint arXiv:2403.04797}, 2024{\natexlab{e}}.

\bibitem[Zhang et~al.(2024{\natexlab{f}})Zhang, Sheng, Zhou, Chen, Zheng, Cai, Song, Tian, R{\'e}, Barrett, et~al.]{zhang2024h2o}
Zhenyu Zhang, Ying Sheng, Tianyi Zhou, Tianlong Chen, Lianmin Zheng, Ruisi Cai, Zhao Song, Yuandong Tian, Christopher R{\'e}, Clark Barrett, et~al.
\newblock H2o: Heavy-hitter oracle for efficient generative inference of large language models.
\newblock \emph{Advances in Neural Information Processing Systems}, 36, 2024{\natexlab{f}}.

\bibitem[Zhang et~al.(2023{\natexlab{c}})Zhang, Fang, Chen, Namazi-Rad, and Wang]{zhang2023large}
Zihan Zhang, Meng Fang, Ling Chen, Mohammad-Reza Namazi-Rad, and Jun Wang.
\newblock How do large language models capture the ever-changing world knowledge? a review of recent advances.
\newblock \emph{arXiv preprint arXiv:2310.07343}, 2023{\natexlab{c}}.

\bibitem[Zhong et~al.(2023)Zhong, Guo, Gao, and Wang]{MemoryBank}
Wanjun Zhong, Lianghong Guo, Qiqi Gao, and Yanlin Wang.
\newblock Memorybank: Enhancing large language models with long-term memory.
\newblock \emph{arXiv preprint arXiv:2305.10250}, 2023.

\bibitem[Zhou et~al.(2023)Zhou, Jiang, Cui, Wang, Xiao, Hou, Cotterell, and Sachan]{zhou2023recurrentgpt}
Wangchunshu Zhou, Yuchen~Eleanor Jiang, Peng Cui, Tiannan Wang, Zhenxin Xiao, Yifan Hou, Ryan Cotterell, and Mrinmaya Sachan.
\newblock Recurrentgpt: Interactive generation of (arbitrarily) long text.
\newblock \emph{arXiv preprint arXiv:2305.13304}, 2023.

\bibitem[Zhou et~al.(2024)Zhou, Ou, Ding, Li, Wu, Wang, Chen, Wang, Xu, Zhang, et~al.]{zhou2024symbolic}
Wangchunshu Zhou, Yixin Ou, Shengwei Ding, Long Li, Jialong Wu, Tiannan Wang, Jiamin Chen, Shuai Wang, Xiaohua Xu, Ningyu Zhang, et~al.
\newblock Symbolic learning enables self-evolving agents.
\newblock \emph{arXiv preprint arXiv:2406.18532}, 2024.

\bibitem[Zhu et~al.(2023)Zhu, Yang, Wang, Song, Wu, Wei, and Li]{zhu2023pose}
Dawei Zhu, Nan Yang, Liang Wang, Yifan Song, Wenhao Wu, Furu Wei, and Sujian Li.
\newblock Pose: Efficient context window extension of llms via positional skip-wise training.
\newblock In \emph{The Twelfth International Conference on Learning Representations}, 2023.

\end{thebibliography}
\bibliographystyle{tmlr}

\clearpage
\appendix

\wy{
\section{Additional Details of Method Classification}
\label{sec:details_of_method}
Here we mainly discuss why we state that saving \paex into knowledge base has $O(n)$ storage complexity and saving \paex into memories has $o(n)$ complexity. We define Storage Complexity as a conceptual measure of how the amount of stored information grows with the number of past experiences (denoted by $n$). For example, suppose the system interacts daily with a user for one year, accumulating ~365 days of conversation ($n=365$). In a knowledge-based method that stores each day’s data separately, the storage may grow linearly with the number of experiences, $O(n)$. If each day’s interaction is stored verbatim, the total storage after 365 days is approximately 365 conversation records. In contrast, memory-based methods might employ a forgetting mechanism. For instance, if the system only stores a summarized version of past interactions, it may reduce each day’s information by a certain factor. As $n$ grows large, the total amount of stored data could become sub-linear, reflecting that older interactions are increasingly compressed or partially discarded. For example, after the first few weeks, only a summary of each earlier conversation might remain. Thus, over a long period, you might not need to store all 365 full conversations; instead, you store a few full recent ones and short summaries of older ones, resulting in less than a strictly linear increase in total storage. 

\section{Challenges of Implementing our Proposed Instantiation} 
\label{sec:challenges_of_proposed_instantiation}
Two major challenges are described below: 

\textbf{Limited Real-World Implementation of Memory-Based Methods}: Existing memory-oriented methods  for GPT-4 level large language models (LLMs) largely are essentially texts rather than true “hidden-space” memory. Methods that have memory in hidden space, such as MemoryLLM~\citep{wang2024memoryllm}, InfiniteAttention~\citep{infini-attention}, Memory$^3$~\citep{memory3} illustrate early attempts in this direction. However, these methods have only been successfully applied to relatively small models (up to a few billion parameters)~\citep{wang2024memoryllm} and often lack open-source implementations~\citep{infini-attention,memory3}, making it difficult to scale and integrate with state-of-the-art large LLMs. In contrast, RAG methods can easily be applied on GPT4-level models. We can (1) Create an LSCS based on small models so that memory and RAG can both be incorporated, but these models may have limited capacities (For instance, RULER [4] shows that Llama-3.1-8B only has 32k effective context window.), which might make it hard to create a truly practical LSCS. (2) Ideally if we have a large memory language model where we can introduce RAG, then the built LSCS can be much stronger but as we said the "large memory language model" currently does not exist. 

\textbf{Lack of Appropriate Long-Term Benchmarks}: Current benchmarks, such as $\infty$Bench~\citep{zhang2024infty} or Loong Bench~\citep{loong}, are still at 200k tokens level, and state-of-the-art models can often solve these tasks by naively fitting all relevant context into their windows (for instance, Gemini-1.5 pro has 1M context window). In contrast, LSCS targets much longer time horizons—akin to a system’s entire lifespan—and we currently lack datasets and benchmarks that reflect these extreme scales. As a result, evaluating and validating the proposed LSCS on real-world or lifespan-scale scenarios is not yet feasible.

Despite these limitations, we view LSCS as a forward-looking framework. As larger memory-equipped models, better open-source implementations, and more extensive benchmarks emerge, the LSCS concept could transition from speculation to practical realization.

Although integrating different categories of methods might be challenging, we would like to mention that a promising direction is to use a model that supports both memory tokens and RAG in a unified manner. For instance, MemoryLLM~\citep{wang2024memoryllm} incorporates up to 12,800 memory tokens per layer and a generation context window of 2,048 tokens. Scaling this approach—e.g., to 96k memory tokens per layer and 32k context window—could enable a model to process vast amounts of stored knowledge within its hidden space. Under such a scenario, RAG techniques could then retrieve and feed external knowledge (e.g., from a notepad-like structure) back into the model, creating a seamless interplay between internally stored representations and external references. While fine-tuning such a system remains a challenge, our view is that this would be an infrequent event. In rare cases where fine-tuning is needed, one could preserve the original training and instruction datasets, mixing them into the new training set to regularize and prevent catastrophic forgetting.
}

\end{document}